\documentclass[sigconf]{acmart}

\usepackage{bm}
\usepackage{amsmath}
\usepackage{amsfonts}
\usepackage{amssymb}
\usepackage{amsthm}
\usepackage{multirow}
\usepackage{caption}
\usepackage{subcaption}
\usepackage{enumitem}
\usepackage{makecell}
\usepackage{graphicx}

\hypersetup{colorlinks=true,urlcolor=blue}
\newcommand{\tabincell}[2]{\begin{tabular}{@{}#1@{}}#2\end{tabular}}
\setlist[itemize,1]{leftmargin=2em}

\setcopyright{none}
\acmConference{WWW'18}{April 23--27, 2018}{Lyon, France.}
\acmYear{2018}
\copyrightyear{2018}
\acmPrice{15.00}

\begin{document}
\title[DKN: Deep Knowledge-Aware Network for News Recommendation]{DKN: Deep Knowledge-Aware Network\\for News Recommendation}

\author{Hongwei Wang$^{1,2}$, Fuzheng Zhang$^2$, Xing Xie$^2$, Minyi Guo$^1$}
\authornote{M. Guo is the corresponding author. This work was partially sponsored by the National Basic Research 973 Program of China under Grant 2015CB352403.}
\affiliation{$^1$Shanghai Jiao Tong University, Shanghai, China}
\affiliation{$^2$Microsoft Research Asia, Beijing, China}
\email{wanghongwei55@gmail.com,{fuzzhang,xingx}@microsoft.com,guo-my@cs.sjtu.edu.cn}

\renewcommand{\shortauthors}{H. Wang et al.}

\begin{abstract}
	Online news recommender systems aim to address the information explosion of news and make personalized recommendation for users.
	In general, news language is highly condensed, full of knowledge entities and common sense.
	However, existing methods are unaware of such external knowledge and cannot fully discover latent knowledge-level connections among news.
	The recommended results for a user are consequently limited to simple patterns and cannot be extended reasonably.
	To solve the above problem, in this paper, we propose a \textit{deep knowledge-aware network} (DKN) that incorporates knowledge graph representation into news recommendation.
	DKN is a content-based deep recommendation framework for click-through rate prediction.
	The key component of DKN is a multi-channel and word-entity-aligned \textit{knowledge-aware convolutional neural network} (KCNN) that fuses semantic-level and knowledge-level representations of news.
	KCNN treats words and entities as multiple channels, and explicitly keeps their alignment relationship during convolution.
	In addition, to address users' diverse interests, we also design an \textit{attention} module in DKN to dynamically aggregate a user's history with respect to current candidate news.
	Through extensive experiments on a real online news platform, we demonstrate that DKN achieves substantial gains over state-of-the-art deep recommendation models.
\end{abstract}


\maketitle

\section{Introduction}
	With the advance of the World Wide Web, people's news reading habits have gradually shifted from traditional media such as newspapers and TV to the Internet.
	Online news websites, such as Google News\footnote{\url{https://news.google.com/}} and Bing News\footnote{\url{https://www.bing.com/news}}, collect news from various sources and provide an aggregate view of news for readers.
	A notorious problem with online news platforms is that the volume of articles can be overwhelming to users.
	To alleviate the impact of information overloading, it is critical to help users target their reading interests and make personalized recommendations \cite{phelan2009using, li2010contextual, liu2010personalized, son2013location, bansal2015content, okura2017embedding}.
	
\begin{figure}[t]
		\centering
  		\includegraphics[width=.4\textwidth]{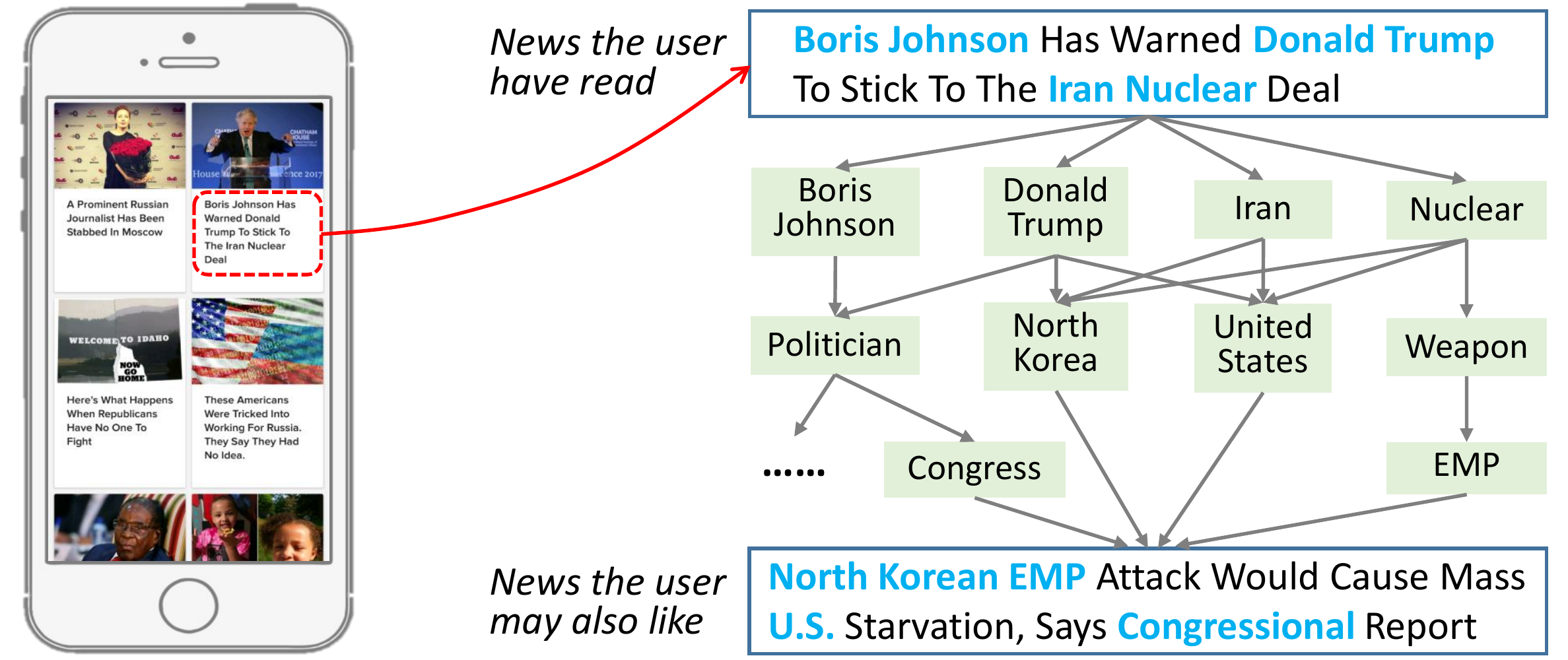}
  		\caption{Illustration of two pieces of news connected through knowledge entities.}
  		\label{fig:buzzfeed}
  		\vspace{-0.1in}
	\end{figure}

	Generally, news recommendation is quite difficult as it poses three major challenges.
	First, unlike other items such as movies \cite{diao2014jointly} and restaurants \cite{fu2014user}, news articles are highly time-sensitive and their relevance expires quickly within a short period (see Section \ref{sec:dd}).
	Out-of-date news are substituted by newer ones frequently, which makes traditional ID-based methods such as collaborative filtering (CF) \cite{wang2011collaborative} less effective.
	Second, people are topic-sensitive in news reading as they are usually interested in multiple specific news categories (see Section \ref{sec:cs}).
	How to dynamically measure a user's interest based on his diversified reading history for current candidate news is key to news recommender systems.
	Third, news language is usually highly condensed and comprised of a large amount of knowledge entities and common sense.
	For example, as shown in Figure \ref{fig:buzzfeed}, a user clicks a piece of news with title ``\textsf{Boris Johnson Has Warned Donald Trump To Stick To The Iran Nuclear Deal}" that contains four knowledge entities: ``\textsf{Boris Johnson}'', ``\textsf{Donald Trump}'', ``\textsf{Iran}'' and ``\textsf{Nuclear}''.
	In fact, the user may also be interested in another piece of news with title ``\textsf{North Korean EMP Attack Would Cause Mass U.S. Starvation, Says Congressional Report}'' with high probability, which shares a great deal of contextual knowledge and is strongly connected with the previous one in terms of commonsense reasoning.
	However, traditional semantic models \cite{mikolov2013distributed} or topic models \cite{blei2003latent} can only find their relatedness based on co-occurrence or clustering structure of words, but are hardly able to discover their latent knowledge-level connection.
	As a result, a user's reading pattern will be narrowed down to a limited circle and cannot be reasonably extended based on existing recommendation methods.
	
	To extract deep logical connections among news, it is necessary to introduce additional \textit{knowledge graph} information into news recommendations.
	A knowledge graph is a type of directed heterogeneous graph in which nodes correspond to \textit{entities} and edges correspond to \textit{relations}.
	Recently, researchers have proposed several academic knowledge graphs such as NELL\footnote{\url{http://rtw.ml.cmu.edu/rtw/}} and DBpedia\footnote{\url{http://wiki.dbpedia.org/}}, as well as commercial ones such as Google Knowledge Graph\footnote{\url{https://www.google.com/intl/bn/insidesearch/features/search/knowledge.html}} and Microsoft Satori\footnote{\url{https://searchengineland.com/library/bing/bing-satori}}.
	These knowledge graphs are successfully applied in scenarios of machine reading\cite{yang2017leveraging}, text classification\cite{wang2017combining}, and word embedding\cite{xu2014rc}.
	
	Considering the above challenges in news recommendation and inspired by the wide success of leveraging knowledge graphs, in this paper, we propose a novel framework that takes advantage of external knowledge for news recommendation, namely the \textit{deep knowledge-aware network} (DKN).
	DKN is a content-based model for click-through rate (CTR) prediction, which takes one piece of candidate news and one user's click history as input, and outputs the probability of the user clicking the news.
	Specifically, for a piece of input news, we first enrich its information by associating each word in the news content with a relevant entity in the knowledge graph.
	We also search and use the set of contextual entities of each entity (i.e., its immediate neighbors in the knowledge graph) to provide more complementary and distinguishable information.
	Then we design a key component in DKN, namely \textit{knowledge-aware convolutional neural networks} (KCNN), to fuse the word-level and knowledge-level representations of news and generate a knowledge-aware embedding vector.
	Distinct from existing work \cite{wang2017combining}, KCNN is:
	1) \textit{multi-channel}, as it treats word embedding, entity embedding, and contextual entity embedding of news as multiple stacked channels just like colored images;
	2) \textit{word-entity-aligned}, as it aligns a word and its associated entity in multiple channels and applies a transformation function to eliminate the heterogeneity of the word embedding and entity embedding spaces.
	
	Using KCNN, we obtain a knowledge-aware representation vector for each piece of news.
	To get a dynamic representation of a user with respect to current candidate news, we use an \textit{attention} module to automatically match candidate news to each piece of clicked news, and aggregate the user's history with different weights.
	The user's embedding and the candidate news' embedding are finally processed by a deep neural network (DNN) for CTR prediction.
	
	Empirically, we apply DKN to a real-world dataset from Bing News with extensive experiments.
	The results show that DKN achieves substantial gains over state-of-the-art deep-learning-based methods for recommendation.
	Specifically, DKN significantly outperforms baselines by $2.8\%$ to $17.0\%$ on F1 and $2.6\%$ to $16.1\%$ on AUC with a significance level of $0.1$.
	The results also prove that the usage of knowledge and an attention module can bring additional $3.5\%$ and $1.4\%$ in improvement, respectively, in the DKN framework.
	Moreover, we present a visualization result of attention values to intuitively demonstrate the efficacy of the usage of the knowledge graph in Section \ref{sec:cs}.

\section{Preliminaries}
	In this section, we present several concepts and models related to this work, including knowledge graph embedding and convolutional neural networks for sentence representation learning.
	
	\subsection{Knowledge Graph Embedding}
	\label{sec:kge}
		A typical knowledge graph consists of millions of entity-relation-entity triples $(h, r, t)$, in which $h$, $r$ and $t$ represent the head, the relation, and the tail of a triple, respectively.
		Given all the triples in a knowledge graph, the goal of knowledge graph embedding is to learn a low-dimensional representation vector for each entity and relation that preserves the structural information of the original knowledge graph.
		Recently, translation-based knowledge graph embedding methods have received great attention due to their concise models and superior performance.
		To be self-contained, we briefly review these translation-based methods in the following.
		
		\begin{itemize}
			\item
			\textbf{TransE} \cite{bordes2013translating} wants $\bf h + \bf r \approx \bf t$ when $(h, r, t)$ holds, where $\bf h$, $\bf r$ and $\bf t$ are the corresponding representation vector of $h$, $r$ and $t$.
			Therefore, TransE assumes the score function
			\begin{equation}
				f_r(h, t) = \| \bf h + \bf r - \bf t \|_2^2
			\end{equation}
			is low if $(h, r, t)$ holds, and high otherwise.
		
			\item
			\textbf{TransH} \cite{wang2014knowledge} allows entities to have different representations when involved in different relations by projecting the entity embeddings into relation hyperplanes:
			\begin{equation}
				f_r(h, t) = \| \bf h_\perp + \bf r - \bf t_\perp \|_2^2,
			\end{equation}
			where ${\bf h}_\perp = {\bf h} - {\bf w}_r^\top {\bf h} {\bf w}_r$ and ${\bf t}_\perp = {\bf t} - {\bf w}_r^\top {\bf t} {\bf w}_r$ are the projections of $\bf h$ and $\bf t$ to the hyperplane ${\bf w}_r$, respectively, and $\| {\bf w}_r \|_2 = 1$.
		
			\item
			\textbf{TransR} \cite{lin2015learning} introduces a projection matrix ${\bf M}_r$ for each relation $r$ to map entity embeddings to the corresponding relation space.
			The score function in TransR is defined as
			\begin{equation}
				f_r(h, t) = \| {\bf h}_r + {\bf r} - {\bf t}_r \|_2^2,
			\end{equation}
			where ${\bf h}_r = {\bf h} {\bf M}_r$ and ${\bf t}_r = {\bf t} {\bf M}_r$.
		
			\item
			\textbf{TransD} \cite{ji2015knowledge} replaces the projection matrix in TransR by the product of two projection vectors of an entity-relation pair:
			\begin{equation}
				f_r(h, t) = \| \bf h_\perp + \bf r - \bf t_\perp \|_2^2,
			\end{equation}
			where ${\bf h}_\perp = ({\bf r}_p {\bf h}_p^\top + {\bf I}) {\bf h}$, ${\bf t}_\perp = ({\bf r}_p {\bf t}_p^\top + {\bf I}) {\bf t}$, ${\bf h}_p$, ${\bf r}_p$ and ${\bf t}_p$ are another set of vectors for entities and relations, and $\bf I$ is the identity matrix.
		\end{itemize}
		
		To encourage the discrimination between correct triples and incorrect triples, for all the methods above, the following margin-based ranking loss is used for training:
		\begin{equation}
			\mathcal L = \sum_{(h, r, t) \in \Delta} \sum_{(h', r, t') \in \Delta'} \max \big( 0, f_r(h, t) + \gamma - f_r(h', t') \big),
		\end{equation}
		where $\gamma$ is the margin, $\Delta$ and $\Delta'$ are the set of correct triples and incorrect triples.

	\subsection{CNN for Sentence Representation Learning}
	\label{sec:cnn_srl}
		\begin{figure}[t]
			\centering
  			\includegraphics[width=.43\textwidth]{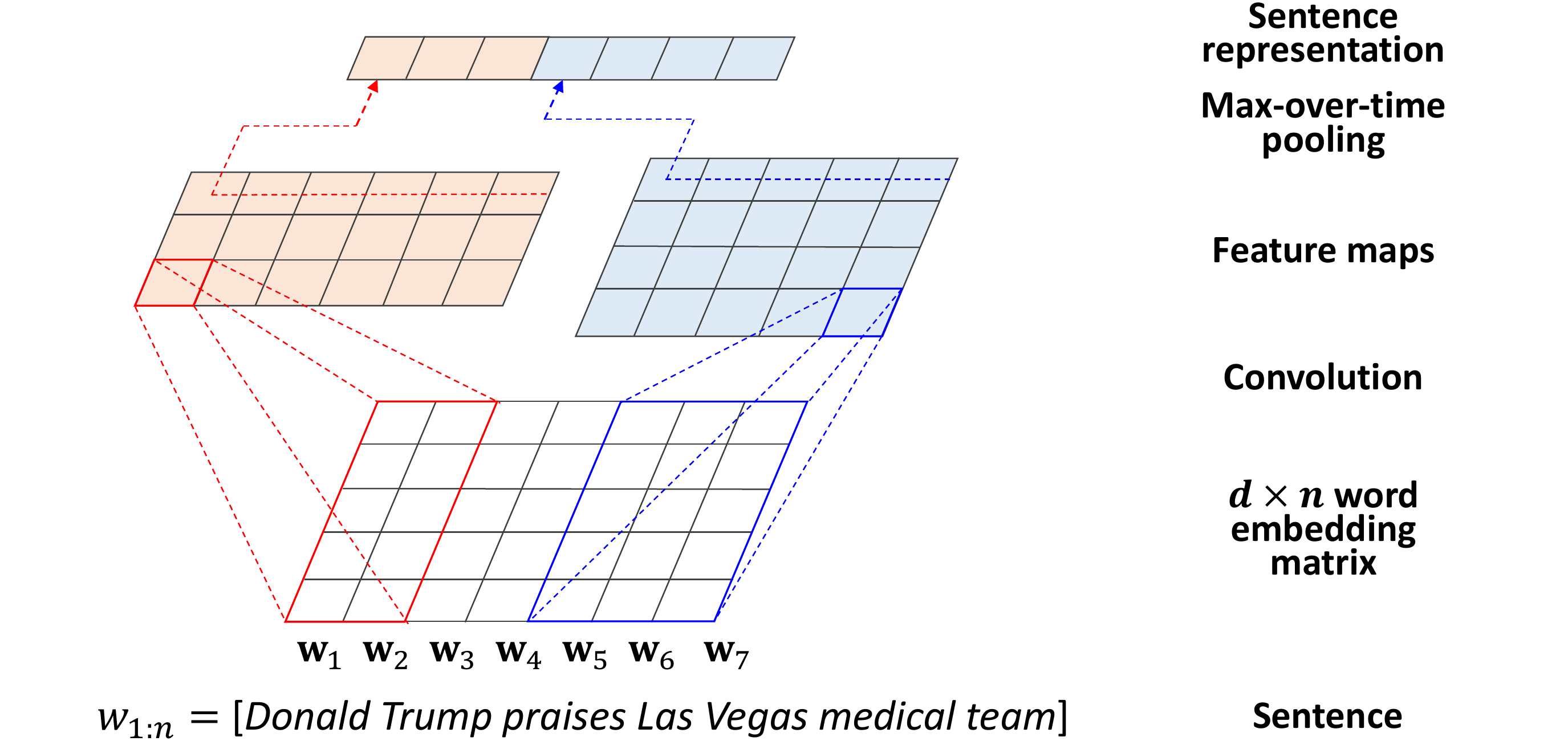}
  			\caption{A typical architecture of CNN for sentence representation learning \cite{kim2014convolutional}.}
  			\label{fig:kim_cnn}
		\end{figure}
		
		Traditional methods \cite{agarwal2009regression, wang2017joint} usually represent sentences using the bag-of-words (BOW) technique, i.e., taking word counting statistics as the feature of sentences.
		However, BOW-based methods ignore word orders in sentences and are vulnerable to the sparsity problem, which leads to poor generalization performance.
		A more effective way to model sentences is to represent each sentence in a given corpus as a distributed low-dimensional vector.
		Recently, inspired by the success of applying convolutional neural networks (CNN) in the filed of computer vision \cite{krizhevsky2012imagenet}, researchers have proposed many CNN-based models for sentence representation learning \cite{kim2014convolutional, kalchbrenner2014convolutional, zhang2015character, conneau2016very} \footnote{Researchers have also proposed other types of neural network models for sentence modeling such as recurrent neural networks \cite{tai2015improved}, recursive neural networks \cite{socher2013recursive}, and hybrid models \cite{lai2015recurrent}. However, CNN-based models are empirically proven to be superior than others \cite{hong2015sentiment}, since they can detect and extract specific local patterns from sentences due to the convolution operation. To keep our presentation focused, we only discuss CNN-based models in this paper.}.
		In this subsection, we introduce a typical type of CNN architecture, namely Kim CNN \cite{kim2014convolutional}.
		
		Figure \ref{fig:kim_cnn} illustrates the architecture of Kim CNN.
		Let $w_{1:n}$ be the raw input of a sentence of length $n$, and ${\bf w}_{1:n} = [{\bf w}_1 \ {\bf w}_2 \ ... \ {\bf w}_n] \in \mathbb R^{d \times n}$ be the word embedding matrix of the input sentence, where ${\bf w}_i \in \mathbb R^{d \times 1}$ is the embedding of the $i$-th word in the sentence and $d$ is the dimension of word embeddings.
		A convolution operation with filter ${\bm h} \in \mathbb R^{d \times l}$ is then applied to the word embedding matrix ${\bf w}_{1:n}$, where $l$ ($l \leq n$) is the window size of the filter.
		Specifically, a feature $c_i$ is generated from a sub-matrix ${\bf w}_{i:i+l-1}$ by
		\begin{equation}
			c_i = f ({\bm h} * {\bf w}_{i:i+l-1} + b),
		\end{equation}
		where $f$ is a non-linear function, $*$ is the convolution operator, and $b \in \mathbb R$ is a bias.
		After applying the filter to every possible position in the word embedding matrix, a feature map
		\begin{equation}
			{\bf c} = [c_1, c_2, ..., c_{n-l+1}]
		\end{equation}
		is obtained, then a max-over-time pooling operation is used on feature map $\bf c$ to identify the most significant feature:
		\begin{equation}
			\tilde c = \max \{ {\bf c} \} = \max \{ c_1, c_2, ..., c_{n-l+1} \}.
		\end{equation}
		One can use multiple filters (with varying window sizes) to obtain multiple features, and these features are concatenated together to form the final sentence representation.

	\section{Problem Formulation}
		We formulate the news recommendation problem in this paper as follows.
		For a given user $i$ in the online news platform, we denote his click history as $\{ t_1^i, t_2^i, ..., t_{N_i}^i \}$, where $t_j^i$ ($j = 1, ..., N_i$) is the title\footnote{In addition to title, it is also viable to use abstracts or snippets of news. In this paper, we only take news titles as input, since a title is a decisive factor affecting users' choice of reading. But note that our approach can be easily generalized to any sort of news-related texts.} of the $j$-th news clicked by user $i$, and $N_i$ is the total number of user $i$'s clicked news.
		Each news title $t$ is composed of a sequence of words, i.e., $t = [w_1, w_2, ...]$, where each word $w$ may be associated with an entity $e$ in the knowledge graph.
		For example, in the title ``\textsf{Trump praises Las Vegas medical team}'', ``\textsf{Trump}'' is linked with the entity ``\textsf{Donald Trump}", while ``\textsf{Las}'' and ``\textsf{Vegas}'' are linked with the entity ``\textsf{Las Vegas}".
		Given users' click history as well as the connection between words in news titles and entities in the knowledge graph, we aim to predict whether user $i$ will click a candidate news $t_j$ that he has not seen before.

\section{Deep Knowledge-Aware Network}
	In this section, we present the proposed DKN model in detail.
	We first introduce the overall framework of DKN, then discuss the process of knowledge distillation from a knowledge graph, the design of knowledge-aware convolutional neural networks (KCNN), and the attention-based user interest extraction, respectively.
	
	\subsection{DKN Framework}
		\begin{figure*}[t]
			\centering
  			\includegraphics[width=.87\textwidth]{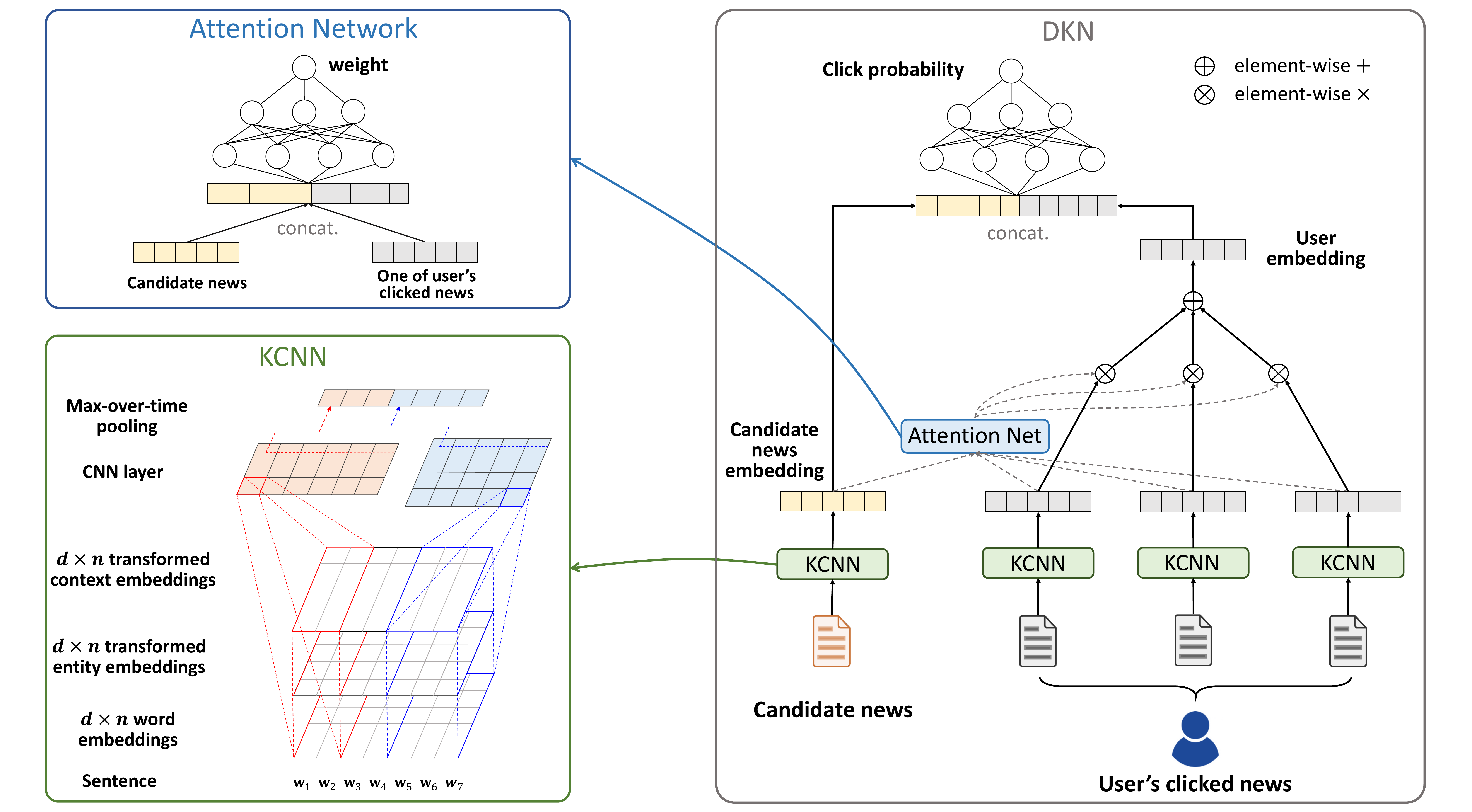}
  			\caption{Illustration of the DKN framework.}
  			\label{fig:framework}
		\end{figure*}
	
		The framework of DKN is illustrated in Figure \ref{fig:framework}.
		We introduce the architecture of DKN from the bottom up.
		As shown in Figure \ref{fig:framework}, DKN takes one piece of candidate news and one piece of a user's clicked news as input.
		For each piece of news, a specially designed KCNN is used to process its title and generate an embedding vector.
		KCNN is an extension of traditional CNN that allows flexibility in incorporating symbolic knowledge from a knowledge graph into sentence representation learning.
		We will detail the process of knowledge distillation in Section \ref{sec:kd} and the KCNN module in Section \ref{sec:kcnn}, respectively.
		By KCNN, we obtain a set of embedding vectors for a user's clicked history.
		To get final embedding of the user with respect to the current candidate news, we use an attention-based method to automatically match the candidate news to each piece of his clicked news, and aggregate the user's historical interests with different weights.
		The details of attention-based user interest extraction are presented in Section \ref{sec:auie}.
		The candidate news embedding and the user embedding are concatenated and fed into a deep neural network (DNN) to calculate the predicted probability that the user will click the candidate news.

	\subsection{Knowledge Distillation}
	\label{sec:kd}
		\begin{figure}[t]
			\centering
  			\includegraphics[width=.4\textwidth]{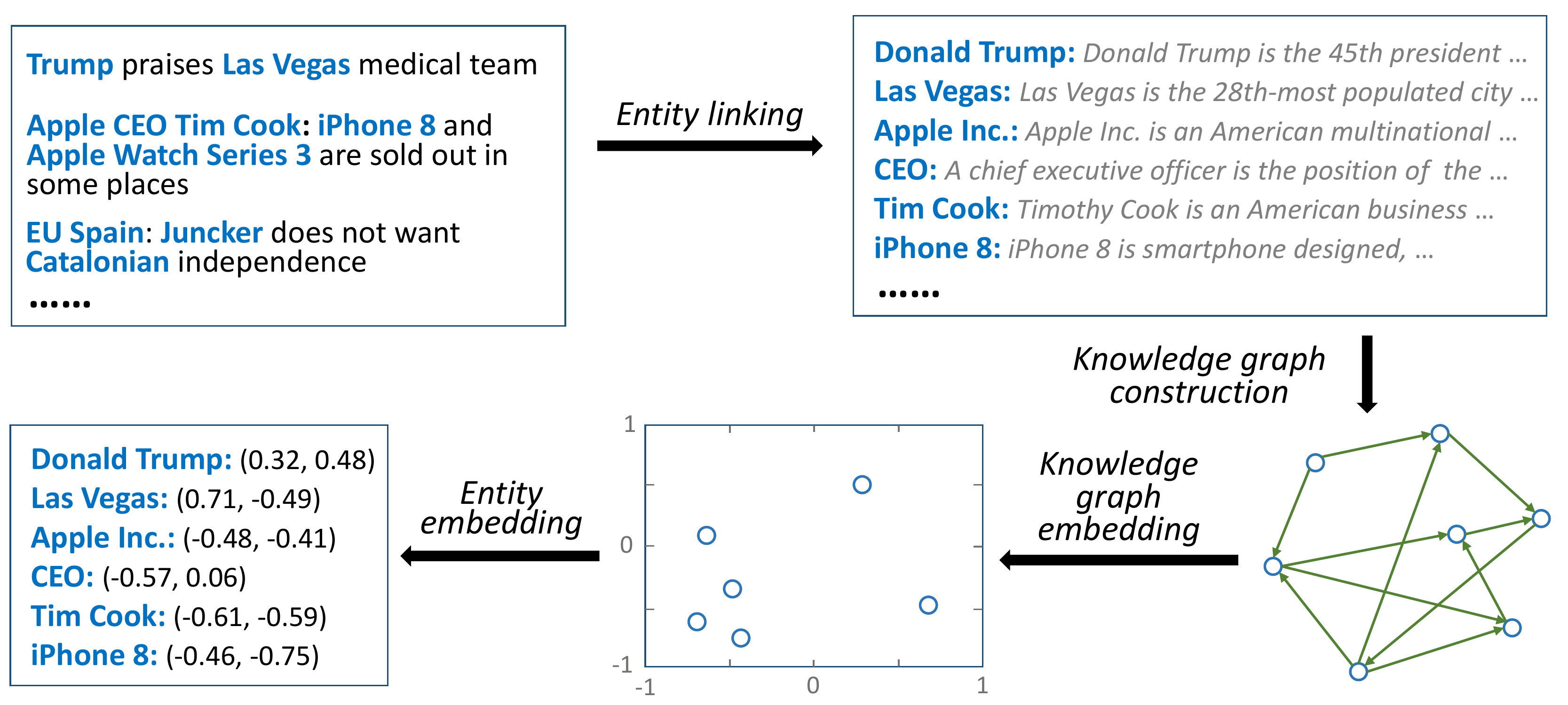}
  			\caption{Illustration of knowledge distillation process.}
  			\label{fig:knowledge_distillation}
		\end{figure}
		
		The process of knowledge distillation is illustrated in Figure \ref{fig:knowledge_distillation}, which consists of four steps.
		First, to distinguish knowledge entities in news content, we utilize the technique of \textit{entity linking} \cite{milne2008learning, sil2013re} to disambiguate mentions in texts by associating them with predefined entities in a knowledge graph.
		Based on these identified entities, we construct a sub-graph and extract all relational links among them from the original knowledge graph.
		Note that the relations among identified entities only may be sparse and lack diversity.
		Therefore, we expand the knowledge sub-graph to all entities within one hop of identified ones.
		Given the extracted knowledge graph, a great many knowledge graph embedding methods, such as TransE \cite{bordes2013translating}, TransH \cite{wang2014knowledge}, TransR \cite{lin2015learning}, and TransD \cite{ji2015knowledge}, can be utilized for entity representation learning.
		Learned entity embeddings are taken as the input for KCNN in the DKN framework.
		
\begin{figure}[t]
			\centering
  			\includegraphics[width=.4\textwidth]{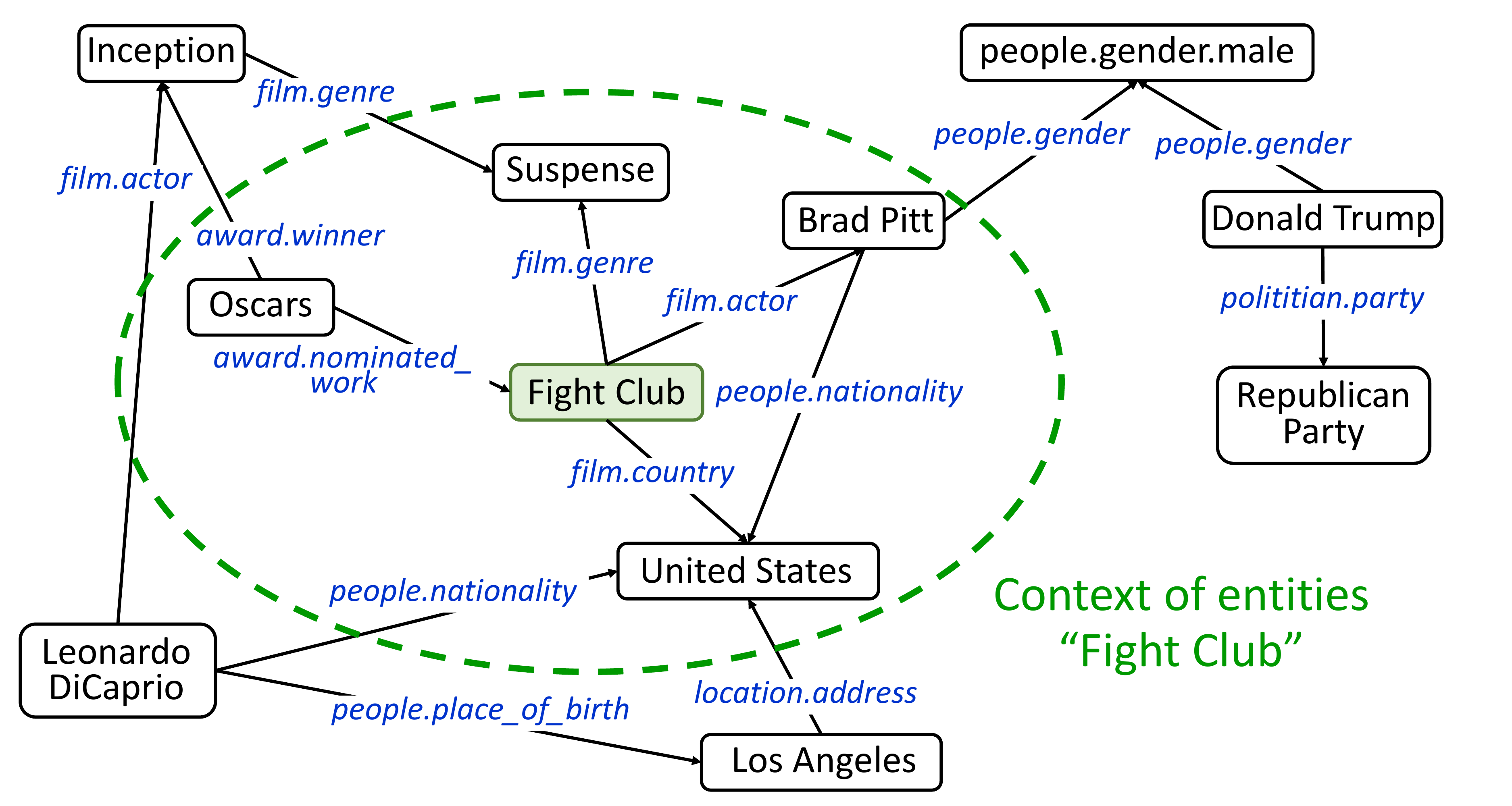}
  			\caption{Illustration of context of an entity in a knowledge graph.}
  			\label{fig:context}
		\end{figure}

		It should be noted that though state-of-the-art knowledge graph embedding methods could generally preserve the structural information in the original graph, we find that the information of learned embedding for a single entity is still limited when used in subsequent recommendations.
		To help identify the position of entities in the knowledge graph, we propose extracting additional contextual information for each entity.
		The ``context'' of entity $e$ is defined as the set of its immediate neighbors in the knowledge graph, i.e.,
		\begin{equation}
			context(e) = \{ e_i \mid (e, r, e_i) \in \mathcal G \ or \ (e_i, r, e) \in \mathcal G \},
		\end{equation}
		where $r$ is a relation and $\mathcal G$ is the knowledge graph.
		Since the contextual entities are usually closely related to the current entity with respect to semantics and logic, the usage of context could provide more complementary information and assist in improving the identifiability of entities.
		Figure \ref{fig:context} illustrates an example of context.
		In addition to use the embedding of ``\textsf{Fight Club}'' itself to represent the entity, we also include its contexts, such as ``\textsf{Suspense}'' (genre), ``\textsf{Brad Pitt}'' (actor), ``\textsf{United States}'' (country) and ``\textsf{Oscars}'' (award), as its identifiers.
		Given the context of entity $e$, the \textit{context embedding} is calculated as the average of its contextual entities:
		\begin{equation}
			\overline {\bf e} = \frac{1}{|context(e)|} \sum_{e_i \in context(e)} {\bf e}_i,
		\end{equation}
		where ${\bf e}_i$ is the \textit{entity embedding} of $e_i$ learned by knowledge graph embedding.
		We empirically demonstrate the efficacy of context embedding in the experiment section.
		
	\subsection{Knowledge-aware CNN}
	\label{sec:kcnn}
		Following the notations used in Section \ref{sec:cnn_srl}, we use $t = w_{1:n} = [w_1, w_2, ..., w_n]$ to denote the raw input sequence of a news title $t$ of length $n$, and ${\bf w}_{1:n} = [{\bf w}_1 \ {\bf w}_2 \ ... \ {\bf w}_n] \in \mathbb R^{d \times n}$ to denote the word embedding matrix of the title, which can be pre-learned from a large corpus or randomly initialized.
		After the knowledge distillation introduced in Section \ref{sec:kd}, each word $w_i$ may also be associated with an entity embedding ${\bf e}_i \in \mathbb R^{k \times 1}$ and the corresponding context embedding $\overline {\bf e}_i  \in \mathbb R^{k \times 1}$, where $k$ is the dimension of entity embedding.
		
		Given the input above, a straightforward way to combine words and associated entities is to treat the entities as ``pseudo words'' and concatenate them to the word sequence \cite{wang2017combining}, i.e.,
		\begin{equation}
			{\bf W} = [{\bf w}_1 \ {\bf w}_2 \ ... \ {\bf w}_n \ {\bf e}_{t_1} \ {\bf e}_{t_2} \ ...],
		\end{equation}
		where $\{ {\bf e}_{t_j} \}$ is the set of entity embeddings associated with this news title.
		The obtained new sentence $\bf W$ is fed into CNN \cite{kim2014convolutional} for further processing.
		However, we argue that this simple concatenating strategy has the following limitations:
		1) The concatenating strategy breaks up the connection between words and associated entities and is unaware of their alignment.
		2) Word embeddings and entity embeddings are learned by different methods, meaning it is not suitable to convolute them together in a single vector space.
		3) The concatenating strategy implicitly forces word embeddings and entity embeddings to have the same dimension, which may not be optimal in practical settings since the optimal dimensions for word  and entity embeddings may differ from each other.
		
		Being aware of the above limitations, we propose a \textit{multi-channel} and \textit{word-entity-aligned} KCNN for combining word semantics and knowledge information.
		The architecture of KCNN is illustrated in the left lower part of Figure \ref{fig:framework}.
		For each news title $t = [w_1, w_2, ..., w_n]$, in addition to use its word embeddings ${\bf w}_{1:n} = [{\bf w}_1 \ {\bf w}_2 \ ... \ {\bf w}_n]$ as input, we also introduce the \textit{transformed entity embeddings}
		\begin{equation}
			g({\bf e}_{1:n}) = [g({\bf e}_1) \ g({\bf e}_2) \ ... \ g({\bf e}_n)]
		\end{equation}
		and \textit{transformed context embeddings}
		\begin{equation}
			g(\overline {\bf e}_{1:n}) = [g(\overline {\bf e}_1) \ g(\overline {\bf e}_2) \ ... \ g(\overline {\bf e}_n)]
		\end{equation}
		as source of input\footnote{${\bf e}_i$ and $\overline {\bf e}_i$ are set as zero if $w_i$ has no corresponding entity.}, where $g$ is the transformation function.
		In KCNN, $g$ can be either linear
		\begin{equation}
			g(\bf e) = M e
		\end{equation}
		or non-linear
		\begin{equation}
		\label{eq:nonlinear}
			g(\bf e) = \tanh(M e + b),
		\end{equation}
		where ${\bf M} \in \mathbb R^{d \times k}$ is the trainable transformation matrix and ${\bf b} \in \mathbb R^{d \times 1}$ is the trainable bias.
		Since the transformation function is continuous, it can map the entity embeddings and context embeddings from the entity space to the word space while preserving their original spatial relationship.
		Note that word embeddings ${\bf w}_{1:n}$, transformed entity embeddings $g({\bf e}_{1:n})$ and transformed context embeddings $g(\overline {\bf e}_{1:n})$ are the same size and serve as the multiple channels analogous to colored images. We therefore align and stack the three embedding matrices as
		\begin{equation}
			{\bf W} = \big[ [{\bf w}_1 \ g({\bf e}_1) \ g(\overline{\bf e}_1)] \ [{\bf w}_2 \ g({\bf e}_2) \ \overline g({\bf e}_2)] \ ... \ [{\bf e}_n \ g({\bf e}_n) \ g(\overline {\bf e}_n)] \big] \in \mathbb R^{d \times n \times 3}.
		\end{equation}
		
		After getting the multi-channel input ${\bf W}$, similar to Kim CNN \cite{kim2014convolutional}, we apply multiple filters ${\bm h} \in \mathbb R^{d \times l \times 3}$ with varying window sizes $l$ to extract specific local patterns in the news title.
		The local activation of sub-matrix ${\bf W}_{i:i+l-1}$ with respect to $\bm h$ can be written as
		\begin{equation}
			c_i^h = f ({\bm h} * {\bf W}_{i:i+l-1} + b),
		\end{equation}
		and we use a max-over-time pooling operation on the output feature map to choose the largest feature:
		\begin{equation}
			\tilde c^h = \max \{ c_1^h, c_2^h, ..., c_{n-l+1}^h \}.
		\end{equation}
		All features $\tilde c^{h_i}$ are concatenated together and taken as the final representation ${\bf e}(t)$ of the input news title $t$, i.e.,
		\begin{equation}
			{\bf e}(t) = [\tilde c^{h_1} \ \tilde c^{h_2} \ ... \ \tilde c^{h_m}],
		\end{equation}
		where $m$ is the number of filters.

	\subsection{Attention-based User Interest Extraction}
	\label{sec:auie}
		Given user $i$ with clicked history $\{ t_1^i, t_2^i, ..., t_{N_i}^i \}$, the embeddings of his clicked news can be written as ${\bf e}(t_1^i)$, ${\bf e}(t_2^i)$, ..., ${\bf e}(t_{N_i}^i)$.
		To represent user $i$ for the current candidate news $t_j$, one can simply average all the embeddings of his clicked news titles:
		\begin{equation}
			{\bf e}(i) = \frac{1}{N_i} \sum_{k=1}^{N_i} {\bf e}(t_k^i).
		\end{equation}
		However, as discussed in the introduction, a user's interest in news topics may be various, and user $i$'s clicked items are supposed to have different impacts on the candidate news $t_j$ when considering whether user $i$ will click $t_j$.
		To characterize user's diverse interests, we use an attention network \cite{wang2017dynamic, zhou2017deep} to model the different impacts of the user's clicked news on the candidate news.
		The attention network is illustrated in the left upper part of Figure \ref{fig:framework}.
		Specifically, for user $i$'s clicked news $t_k^i$ and candidate news $t_j$, we first concatenate their embeddings, then apply a DNN $\mathcal H$ as the attention network and the softmax function to calculate the normalized impact weight: 
		\begin{equation}
			s_{t_k^i, t_j} = \ \text{softmax} \Big( \mathcal H \big( {\bf e}(t_k^i), {\bf e}(t_j) \big) \Big) = \frac{\exp \Big( \mathcal H \big( {\bf e}(t_k^i), {\bf e}(t_j) \big) \Big)}{\sum_{k=1}^{N_i} \exp \Big( \mathcal H \big( {\bf e}(t_k^i), {\bf e}(t_j) \big) \Big)}.
		\end{equation}
		The attention network $\mathcal H$ receives embeddings of two news titles as input and outputs the impact weight.
		The embedding of user $i$ with respect to the candidate news $t_j$ can thus be calculated as the weighted sum of his clicked news title embeddings:
		\begin{equation}
			{\bf e}(i) = \sum_{k=1}^{N_i} s_{t_k^i, t_j} {\bf e}(t_k^i).
		\end{equation}
		
		Finally, given user $i$'s embedding ${\bf e}(i)$ and candidate news $t_j$'s embedding ${\bf e}(t_j)$, the probability of user $i$ clicking news $t_j$ is predicted by another DNN $\mathcal G$:
		\begin{equation}
			p_{i, t_j} = \mathcal G \big( {\bf e}(i), {\bf e}(t_j) \big).
		\end{equation}
		
		We will demonstrate the efficacy of the attention network in the experiment section.

\section{Experiments}
	In this section, we present our experiments and the corresponding results, including dataset analysis and comparison of models.
	We also give a case study about user's reading interests and make discussions on tuning hyper-parameters.
	
	\subsection{Dataset Description}
	\label{sec:dd}			
\begin{figure}
			\centering
            		\begin{subfigure}[b]{0.23\textwidth}
                		\includegraphics[width=\textwidth]{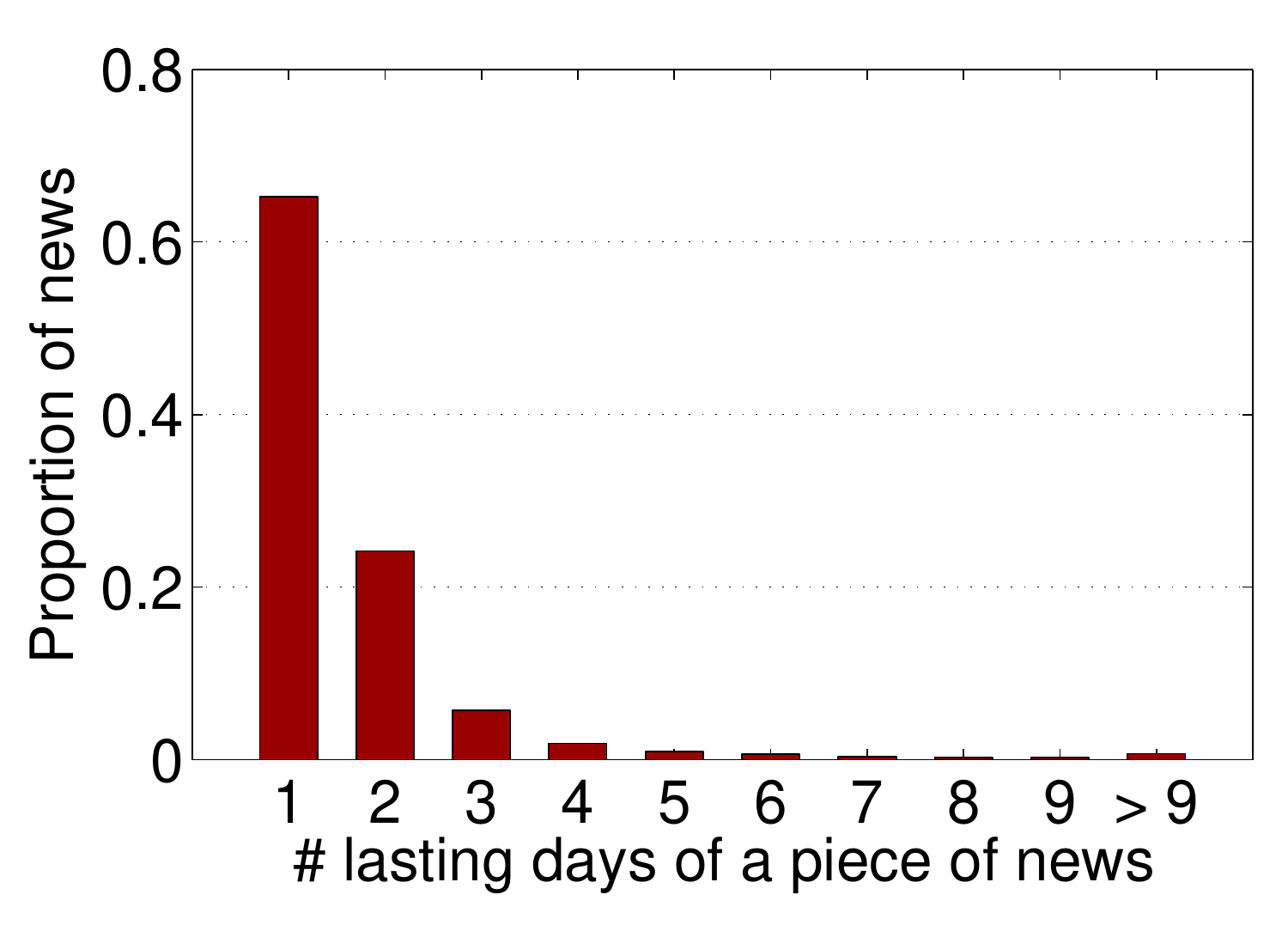}
                		\caption{Distribution of the length of news life cycle}
                		\label{fig:statistics_a}
            		\end{subfigure}
            		\hfill
            		\begin{subfigure}[b]{0.23\textwidth}
                		\includegraphics[width=\textwidth]{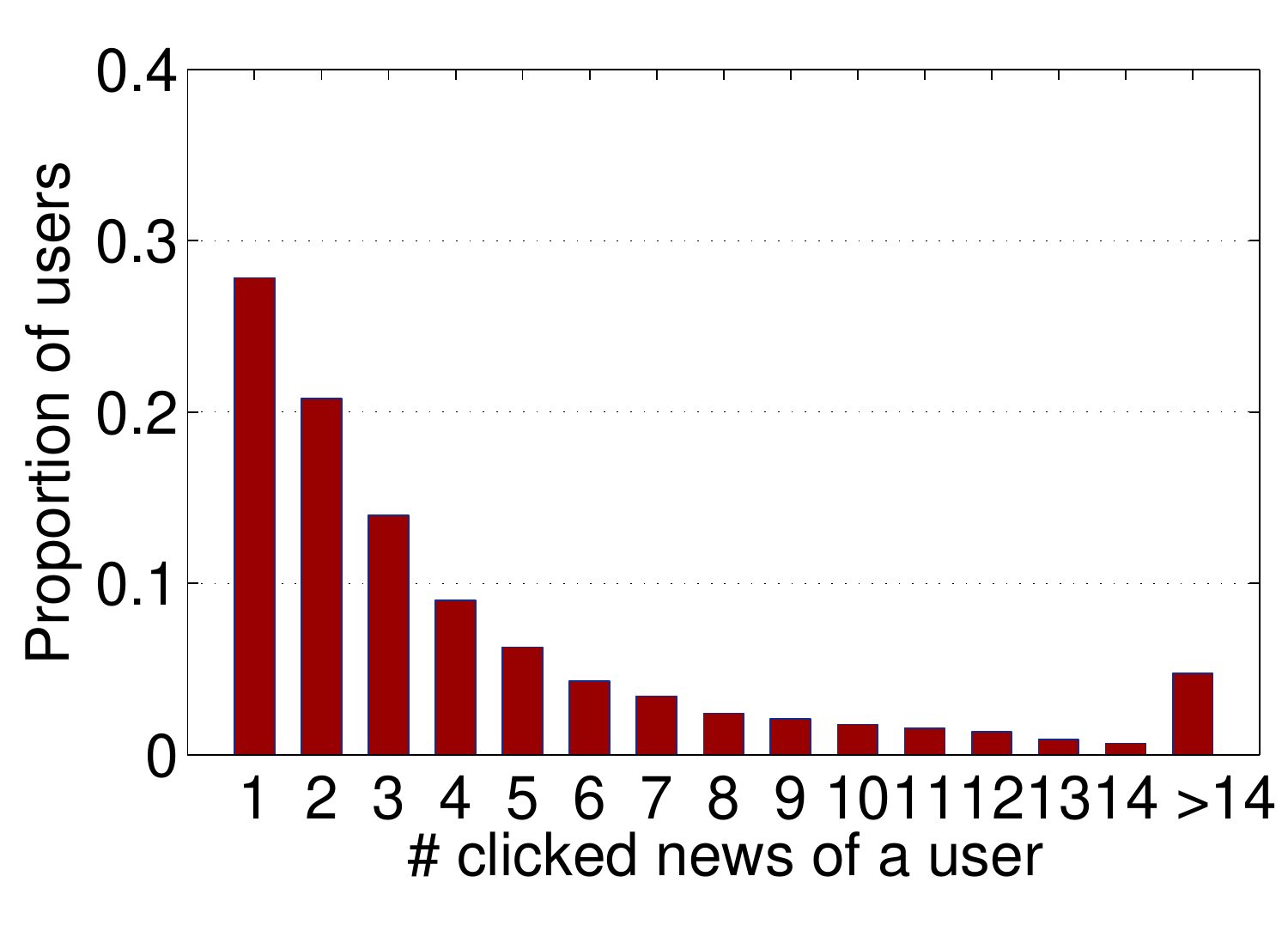}
                		\caption{Distribution of the number of clicked news of a user}
                		\label{fig:statistics_b}
            		\end{subfigure}
            		\hfill
            		\begin{subfigure}[b]{0.23\textwidth}
                		\includegraphics[width=\textwidth]{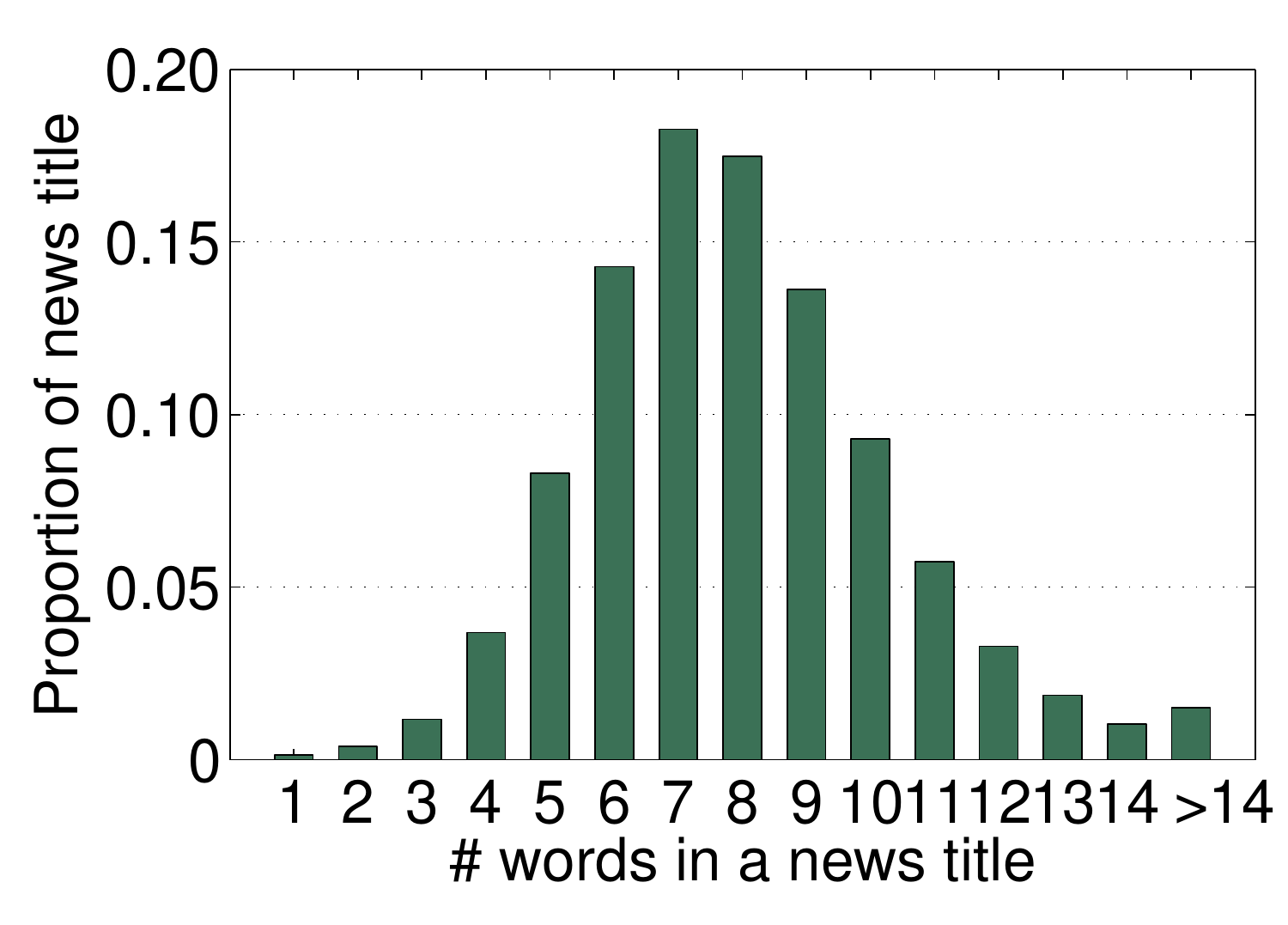}
                		\caption{Distribution of the number of words in a news title}
                		\label{fig:statistics_c}
            		\end{subfigure}
        		\hfill
            		\begin{subfigure}[b]{0.23\textwidth}
                		\includegraphics[width=\textwidth]{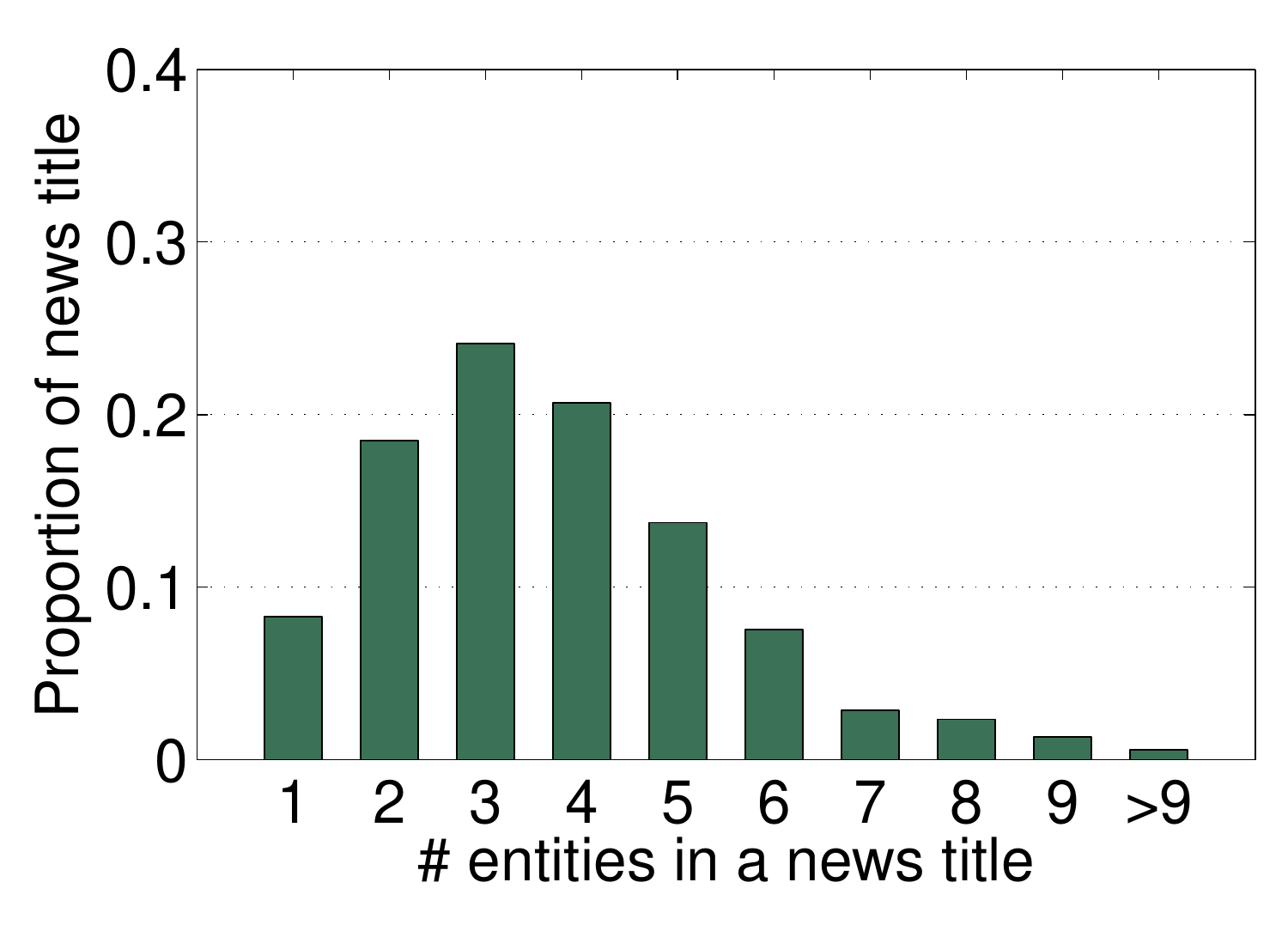}
                		\caption{Distribution of the number of entities in a news title}
                		\label{fig:statistics_d}
            		\end{subfigure}
            		\hfill
            		\begin{subfigure}[b]{0.23\textwidth}
                		\includegraphics[width=\textwidth]{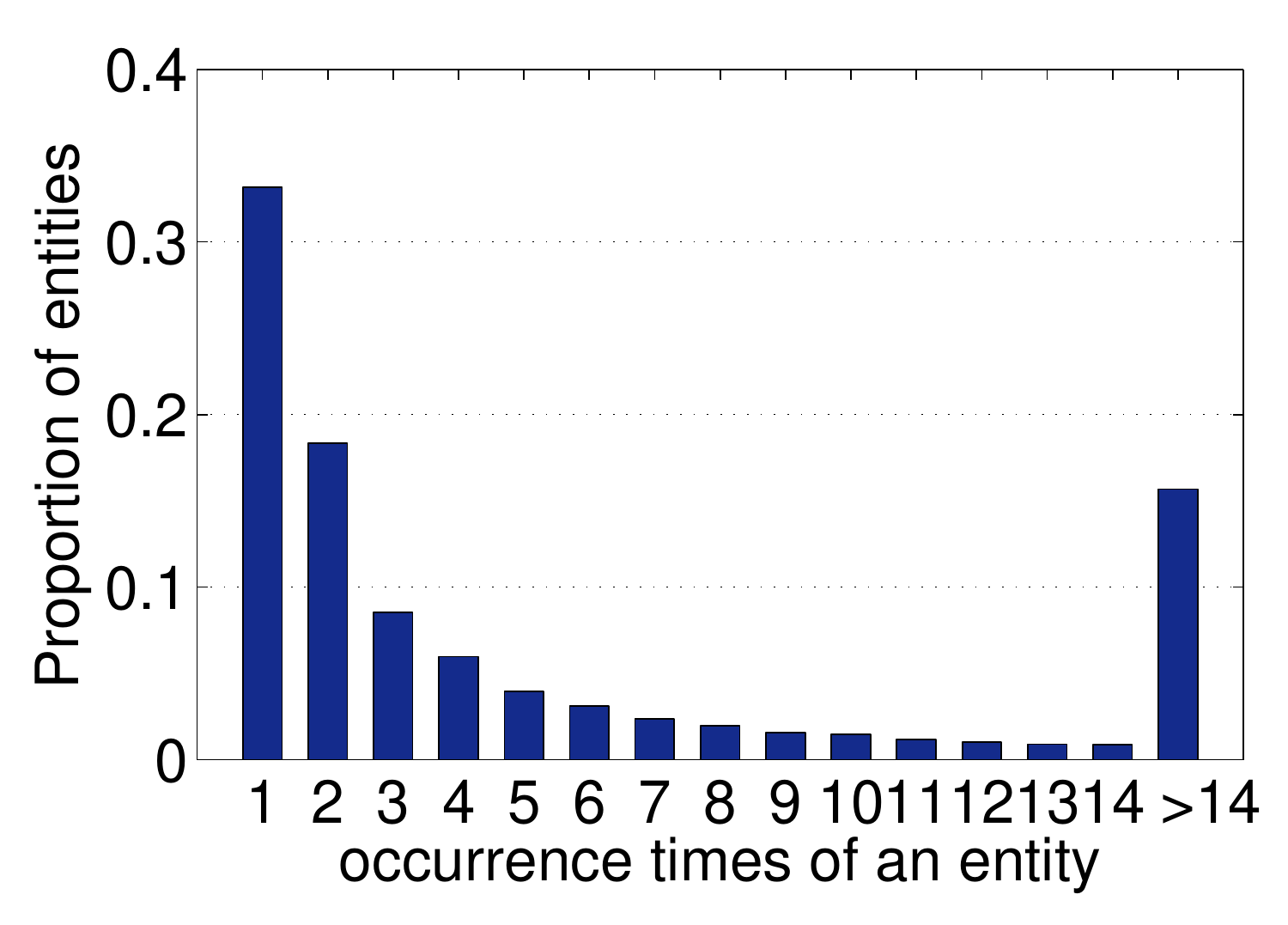}
                		\caption{Distribution of the occurrence times of an entity in the news dataset}
                		\label{fig:statistics_e}
            		\end{subfigure}
            		\hfill
            		\begin{subfigure}[b]{0.23\textwidth}
                		\includegraphics[width=\textwidth]{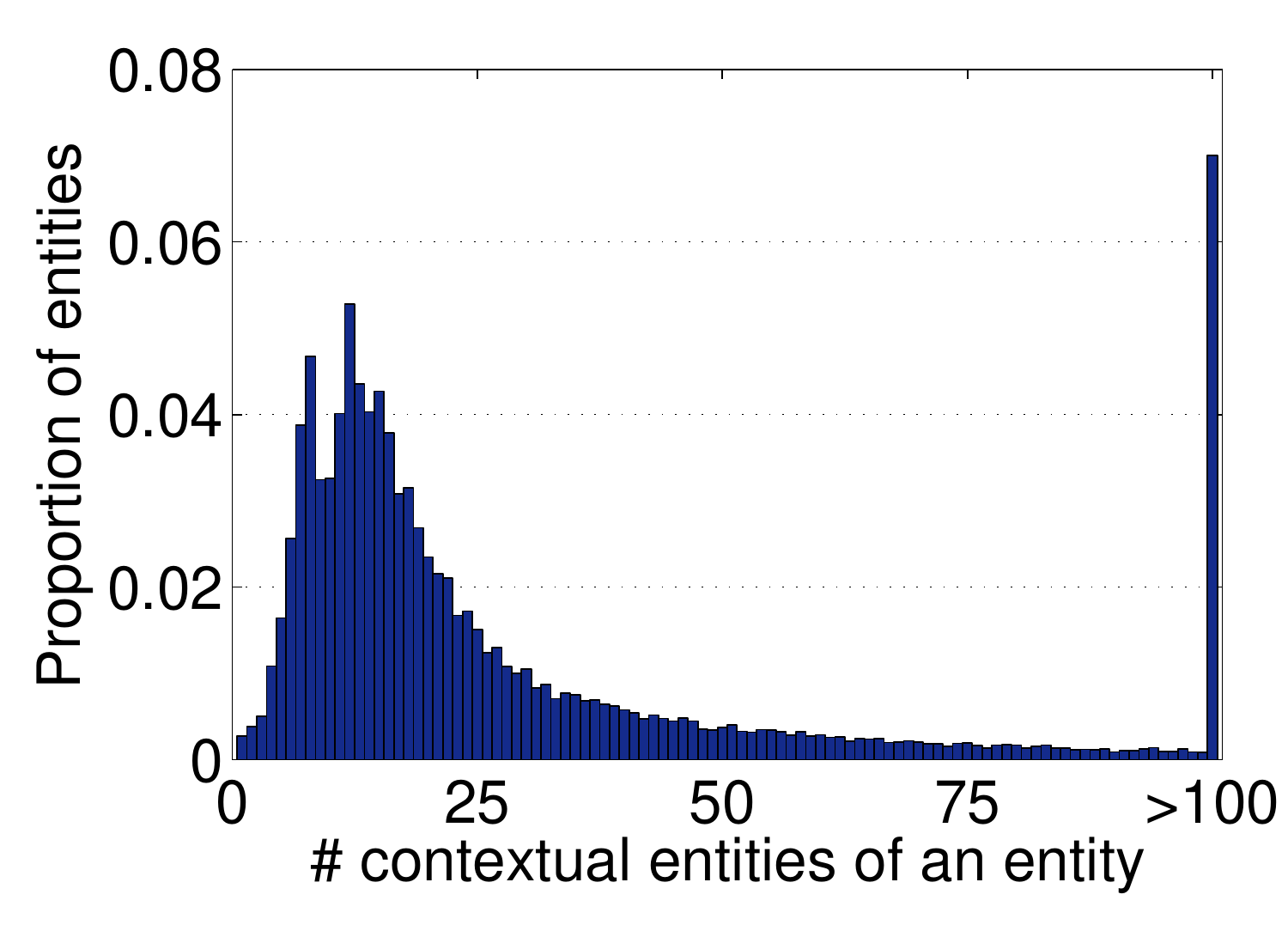}
                		\caption{Distribution of the number of contextual entities of an entity in the knowledge graph}
                		\label{fig:statistics_f}
            		\end{subfigure}
            		\caption{Illustration of statistical distributions in news dataset and extracted knowledge graph.}
            		\label{fig:statistics}
        	\end{figure}

		Our dataset comes from the server logs of Bing News.
		Each piece of log mainly contains the timestamp, user id, news url, news title, and click count (0 for no click and 1 for click).
		We collect a randomly sampled and balanced dataset from October 16, 2016 to June 11, 2017 as the training set, and from June 12, 2017 to August 11, 2017 as the test set.
		Additionally, we search all occurred entities in the dataset as well as the ones within their one hop in the Microsoft Satori knowledge graph, and extract all edges (triples) among them with confidence greater than 0.8.
		The basic statistics and distributions of the news dataset and the extracted knowledge graph are shown in Table \ref {table:statistics} and Figure \ref{fig:statistics}, respectively.
		
		\begin{table}
			\centering
			\small
			\caption{Basic statistics of the news dataset and the extracted knowledge graph.}
			\vspace{-0.1in}
			\begin{tabular}{|c|r||c|r|}
				\hline
				\# users & 141,487 & \# triples & 7,145,776\\
				\hline
				\# news & 535,145 & avg. \# words per title & 7.9\\
				\hline
				 \# logs & 1,025,192 & avg. \# entities per title & 3.7\\
				\hline
				\# entities & 336,350  & \multirow{2}{*}{\tabincell{l}{avg. \# contextual\\entities per entity}} & \multirow{2}{*}{42.5}\\
				\cline{1-2}
				\# relations & 4,668  &  &\\
				\hline
			\end{tabular}
			\label{table:statistics}
			\scriptsize \flushleft{``\#'' denotes ``the number of''.}
		\end{table}
		
		Figure \ref{fig:statistics_a} illustrates the distribution of the length of the news life cycle, where we define the life cycle of a piece of news as the period from its publication date to the date of its last received click.
		We observe that about $90\%$ of news are clicked within two days, which proves that online news is extremely time-sensitive and are substituted by newer ones with high frequency.
		Figure \ref{fig:statistics_b} illustrates the distribution of the number of clicked pieces of news for a user.
		$77.9\%$ of users clicked no more than five pieces of news, which demonstrates the data sparsity in the news recommendation scenario.
		
		Figures \ref{fig:statistics_c} and \ref{fig:statistics_d} illustrate the distributions of the number of words (without stop words) and entities in a news title, respectively.
		The average number per title is $7.9$ for words and $3.7$ for entities, showing that there is one entity in almost every two words in news titles on average.
		The high density of the occurrence of entities also empirically justifies the design of KCNN.
		
		Figures \ref{fig:statistics_e} and \ref{fig:statistics_f} present the distribution of occurrence times of an entity in the news dataset and the distribution of the number of contextual entities of an entity in extracted knowledge graph, respectively.
		We can conclude from the two figures that the occurrence pattern of entities in online news is sparse and has a long tail ($80.4\%$ of entities occur no more than ten times), but entities generally have abundant contexts in the knowledge graph: the average number of context entities per entity is $42.5$ and the maximum is $140,737$.
		Therefore, contextual entities can greatly enrich the representations for a single entity in news recommendation.
		
	\subsection{Baselines}
		We use the following state-of-the-art methods as baselines in our experiments:
		\begin{itemize}
			\item
				\textbf{LibFM} \cite{rendle2012factorization} is a state-of-the-art feature-based factorization model and widely used in CTR scenarios.
				In this paper, the input feature of each piece of news for LibFM is comprised of two parts: TF-IDF features and averaged entity embeddings.
				We concatenate the feature of a user and candidate news to feed into LibFM.
			\item
				\textbf{KPCNN} \cite{wang2017combining} attaches the contained entities to the word sequence of a news title and uses Kim CNN to learn representations of news, as introduced in Section \ref{sec:kcnn}.
			\item
				\textbf{DSSM} \cite{huang2013learning} is a deep structured semantic model for document ranking using word hashing and multiple fully-connected layers.
				In this paper, the user's clicked news is treated as the query and the candidate news are treated as the documents.
			\item
				\textbf{DeepWide} \cite{cheng2016wide} is a general deep model for recommendation, combining a (wide) linear channel with a (deep) non-linear channel.
				Similar to LibFM, we use the concatenated TF-IDF features and averaged entity embeddings as input to feed both channels.
			\item
				\textbf{DeepFM} \cite{guo2017deepfm} is also a general deep model for recommendation, which combines a component of factorization machines and a component of deep neural networks that share the input.
				We use the same input as in LibFM for DeepFM.
			\item
				\textbf{YouTubeNet} \cite{covington2016deep} is proposed to recommend videos from a large-scale candidate set in YouTube using a deep candidate generation network and a deep ranking network.
				In this paper, we adapt the deep raking network to the news recommendation scenario.
			\item
				\textbf{DMF} \cite{xue2017deep} is a deep matrix factorization model for recommender systems which uses multiple non-linear layers to process raw rating vectors of users and items.
				We ignore the content of news and take the implicit feedback as input for DMF.
		\end{itemize}
		
		Note that except for LibFM, other baselines are all based on deep neural networks since we aim to compare our approach with state-of-the-art deep learning models.
		Additionally, except for DMF which is based on collaborative filtering, other baselines are all content-based or hybrid methods.

	\subsection{Experiment Setup}
	\label{sec:es}
		We choose TransD \cite{ji2015knowledge} to process the knowledge graph and learn entity embeddings, and use the non-linear transformation function in Eq. (\ref{eq:nonlinear}) in KCNN.
		The dimension of both word embeddings and entity embeddings are set as $100$.
		The number of filters are set as $100$ for each of the window sizes $1$, $2$, $3$, $4$.		
		We use Adam \cite{kingma2014adam} to train DKN by optimizing the log loss.
		We will further study the variants of DKN and the sensitivity of key parameters in Sections \ref{sec:cm} and \ref{sec:ps}, respectively.
		To compare DKN with baselines, we use \textit{F1} and \textit{AUC} value as the evaluation metrics.
		
		The key parameter settings for baselines are as follows.
		For KPCNN, the dimensions of word embeddings and entity embeddings are both set as $100$.
		For DSSM, the dimension of semantic feature is set as $100$.
		For DeepWide, the final representations for deep and wide components are both set as $100$.
		For YouTubeNet, the dimension of final layer is set as $100$.
		For LibFM and DeepFM, the dimensionality of the factorization machine is set as $\{1, 1, 0 \}$.
		For DMF, the dimension of latent representation for users and items is set as $100$.
		The above settings are for fair consideration.
		Other parameters in the baselines are set as default.
		Each experiment is repeated five times, and we report the average and maximum deviation as results.

		\begin{table}[t]
			\centering
			\small
			\caption{Comparison of different models.}
			\vspace{-0.1in}
			\begin{tabular}{l|l|l|c}
				\hline
				\makecell[c]{Models$^*$} & \makecell[c]{F1} & \makecell[c]{AUC} & \textit{p}-value$^{**}$\\
				\hline
				DKN & \textbf{68.9 $\pm$ 1.5} & \textbf{65.9 $\pm$ 1.2} & $-$ \\
				LibFM & 61.8 $\pm$ 2.1 (-10.3$\%$) & 59.7 $\pm$ 1.8 (-9.4$\%$) & $<10^{-3}$ \\
				LibFM(-) & 61.1 $\pm$ 1.9 (-11.3$\%$) & 58.9 $\pm$ 1.7 (-10.6$\%$) & $<10^{-3}$ \\
				KPCNN & 67.0 $\pm$ 1.6 (-2.8$\%$) & 64.2 $\pm$ 1.4 (-2.6$\%$) & 0.098 \\
				KPCNN(-) & 65.8 $\pm$ 1.4 (-4.5$\%$) & 63.1 $\pm$ 1.5 (-4.2$\%$) & 0.036 \\
				DSSM & 66.7 $\pm$ 1.8 (-3.2$\%$) & 63.6 $\pm$ 2.0 (-3.5$\%$) & 0.063 \\
				DSSM(-) & 66.1 $\pm$ 1.6 (-4.1$\%$) & 63.2 $\pm$ 1.8 (-4.1$\%$) & 0.045 \\
				DeepWide & 66.0 $\pm$1.2 (-4.2$\%$) & 63.3 $\pm$ 1.5 (-3.9$\%$) & 0.039 \\
				DeepWide(-) & 63.7 $\pm$ 0.9 (-7.5$\%$) & 61.5 $\pm$ 1.1 (-6.7$\%$) & 0.004 \\
				DeepFM & 63.8 $\pm$ 1.5 (-7.4$\%$) & 61.2 $\pm$ 2.3 (-7.1$\%$) & 0.014 \\
				DeepFM(-) & 64.0 $\pm$ 1.9 (-7.1$\%$) & 61.1 $\pm$ 1.8 (-7.3$\%$) & 0.007 \\
				YouTubeNet & 65.5 $\pm$ 1.2 (-4.9$\%$) & 63.0 $\pm$ 1.4 (-4.4$\%$) & 0.025 \\
				YouTubeNet(-) & 65.1 $\pm$ 0.7 (-5.5$\%$) & 62.1 $\pm$ 1.3 (-5.8$\%$) & 0.011 \\
				DMF & 57.2 $\pm$ 1.2 (-17.0$\%$) & 55.3 $\pm$ 1.0 (-16.1$\%$) & $<10^{-3}$ \\
				\hline
			\end{tabular}
			\label{table:comparison}
			\scriptsize \flushleft{* ``(-)'' denotes ``without input of entity embeddings''.}
			\scriptsize \flushleft{** $p$-value is the probability of no significant difference with DKN on AUC by \textit{t}-test.}
		\end{table}
			
		\begin{figure}[t]
			\centering
  			\includegraphics[width=.38\textwidth]{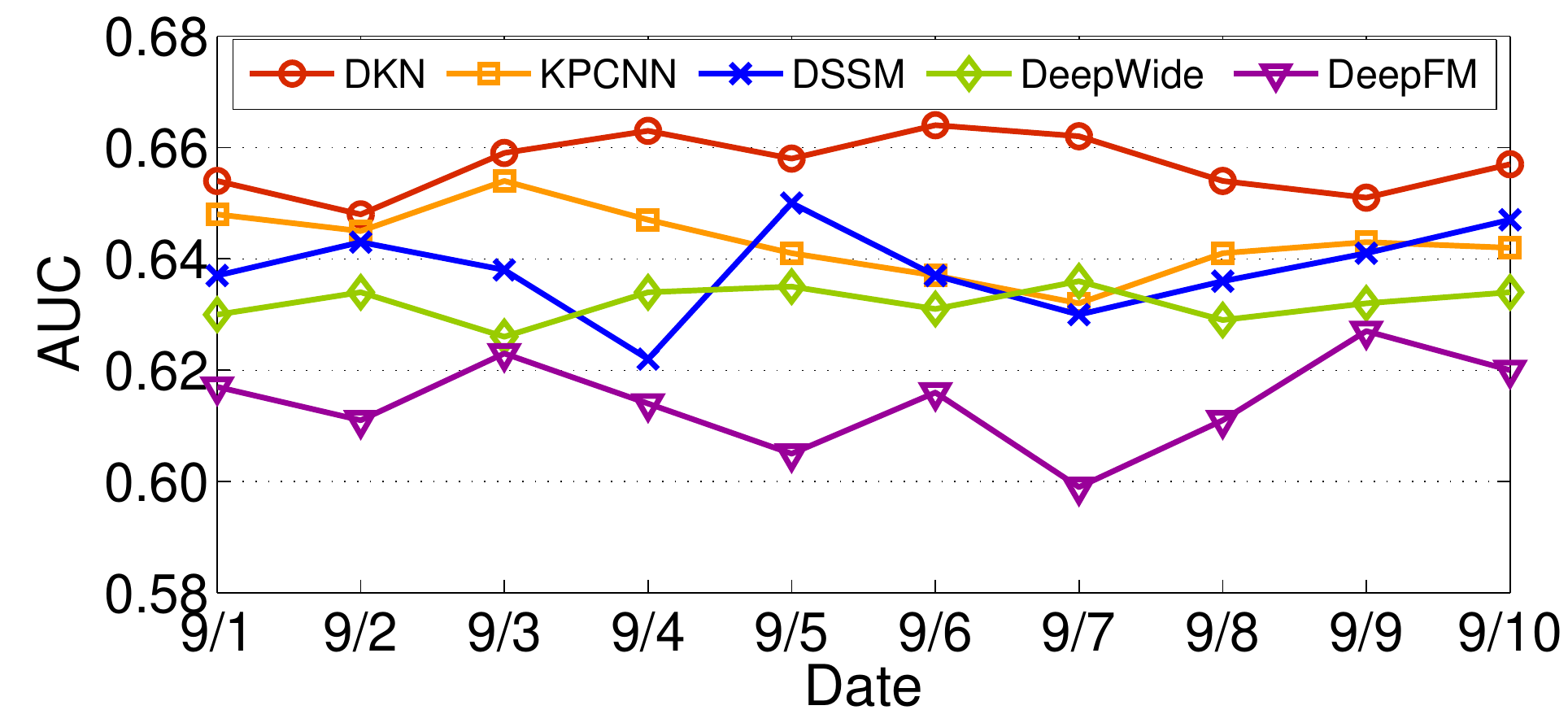}
  			\caption{AUC score of DKN and baselines over ten days (Sep. 01-10, 2017).}
  			\label{fig:result}
		\end{figure}
			
		\begin{table}[t]
			\centering
			\small
			\caption{Comparison among DKN variants.}
			\vspace{-0.1in}
			\begin{tabular}{l|l|l}
				\hline
				\makecell[c]{Variants} & \makecell[c]{F1} & \makecell[c]{AUC}\\
				\hline
				DKN with entity and context emd. & \textbf{68.8 $\pm$ 1.4} & \textbf{65.7 $\pm$ 1.1}\\
				DKN with entity emd. only & 67.2 $\pm$ 1.2 & 64.8 $\pm$ 1.0\\
				DKN with context emd. only & 66.5 $\pm$ 1.5 & 64.2 $\pm$ 1.3\\
				DKN without entity nor context emd. & 66.1 $\pm$1.4 & 63.5 $\pm$ 1.1\\
				\hline
				DKN + TransE & 67.6 $\pm$ 1.6 & 65.0 $\pm$ 1.3\\
				DKN + TransH & 67.3 $\pm$ 1.3 & 64.7 $\pm$ 1.2 \\
				DKN + TransR & 67.9 $\pm$ 1.5 & 65.1 $\pm$ 1.5\\
				DKN + TransD & \textbf{68.8 $\pm$ 1.3} & \textbf{65.8 $\pm$ 1.4}\\
				\hline
				DKN with non-linear mapping & \textbf{69.0 $\pm$ 1.7} & \textbf{66.1 $\pm$ 1.4}\\
				DKN with linear mapping & 67.1 $\pm$ 1.5 & 64.9 $\pm$ 1.3\\
				DKN without mapping & 66.7 $\pm$ 1.6 & 63.7 $\pm$ 1.6\\
				\hline
				DKN with attention & \textbf{68.7 $\pm$ 1.3} & \textbf{65.7 $\pm$ 1.2}\\
				DKN without attention & 67.0 $\pm$ 1.0 & 64.8 $\pm$ 0.8\\
				\hline
			\end{tabular}
			\label{table:variants}
		\end{table}
		
		\begin{table*}
			\centering
			\small
			\setlength{\tabcolsep}{4pt}
			\caption{Illustration of training and test logs for a randomly sampled user (training logs with label 0 are omitted).}
			\vspace{-0.1in}
			\begin{tabular}{c|c|c|l|l|c||c}
				\hline
				& \textbf{No.} & \textbf{Date} & \makecell[c]{\textbf{News title}} & \makecell[c]{\textbf{Entities}} & \textbf{Label} & \textbf{Category}\\
				\hline
				\multirow{9}{*}{\rotatebox{90}{training}} & 1 & 12/25/2016 & Elon Musk teases huge upgrades for Tesla's supercharger network & Elon Musk; Tesla Inc. & 1 & Cars\\
				& 2 & 03/25/2017 & Elon Musk offers Tesla Model 3 sneak peek & Elon Musk; Tesla Model 3 & 1 & Cars\\
				& 3 & 12/14/2016 & Google fumbles while Tesla sprints toward a driverless future & Google Inc.; Tesla Inc. & 1 & Cars\\
				& 4 & 12/15/2016 & Trump pledges aid to Silicon Valley during tech meeting & Donald Trump; Silicon Valley & 1 & Politics\\
				& 5 & 03/26/2017 & Donald Trump is a big reason why the GOP kept the Montana House seat & Donald Trump; GOP; Montana & 1 & Politics\\
				& 6 & 05/03/2017 & North Korea threat: Kim could use nuclear weapons as ``blackmail'' & North Korea; Kim Jong-un & 1 & Politics\\
				& 7 & 12/22/2016 & Microsoft sells out of unlocked Lumia 950 and Lumia 950 XL in the US & Microsoft; Lumia; United States & 1 & Other\\
				& 8 & 12/08/2017 & 6.5 magnitude earthquake recorded off the coast of California & earthquake; California & 1 & Other\\
				& & & ...... & & &\\
				\hline
				\multirow{4}{*}{\rotatebox{90}{test}} & 1 & 07/08/2017 & Tesla makes its first Model 3 & Tesla Inc; Tesla Model 3 & 1 & Cars\\
				& 2 & 08/13/2017 & General Motors is ramping up its self-driving car: Ford should be nervous & General Motors; Ford Inc. & 1 & Cars\\
				& 3 & 06/21/2017 & Jeh Johnson testifies on Russian interference in 2016 election & Jeh Johnson; Russian & 1 & Politics\\
				& 4 & 07/16/2017 & ``Game of Thrones'' season 7 premiere: how you can watch & Game of Thrones & 0 & Other\\
				\hline
			\end{tabular}
			\label{table:case_study}
		\end{table*}
	
	\subsection{Results}
	\label{sec:cm}
		In this subsection, we present the results of comparison of different models and the comparison among variants of DKN.
	
		\subsubsection{Comparison of different models.}
			The results of comparison of different models are shown in Table \ref{table:comparison}.
			For each baseline in which the input contains entity embedding, we also remove the entity embedding from input to see how its performance changes (denoted by ``(-)'').
			Additionally, we list the improvements of baselines compared with DKN in brackets and calculate the \textit{p}-value of statistical significance by \textit{t}-test.
			Several observations stand out from Table \ref{table:comparison}:
			\begin{itemize}
				\item
					The usage of entity embedding could boost the performance of most baselines.
					For example, the AUC of KPCNN, DeepWide, and YouTubeNet increases by $1.1\%$, $1.8\%$ and $1.1\%$, respectively.
					However, the improvement for DeepFM is less obvious.
					We try different parameter settings for DeepFM and find that if the AUC drops to about $0.6$, the improvement brought by the usage of knowledge could be up to $0.5\%$.
					The results show that FM-based method cannot take advantage of entity embedding stably in news recommendation.
				\item
					DMF performs worst among all methods.
					This is because DMF is a CF-based method, but news is generally highly time-sensitive with a short life cycle.
					The result proves our aforementioned claim that CF methods cannot work well in the news recommendation scenario.
				\item
					Except for DMF, other deep-learning-based baselines outperform LibFM by $2.0\%$ to $5.2\%$ on F1 and by $1.5\%$ to $4.5\%$ on AUC, which suggests that deep models are effective in capturing the non-linear relations and dependencies in news data.
				\item
					The architecture of DeepWide and YouTubeNet is similar in the news recommendation scenario, thus we can observe comparable performance of the two methods.
					DSSM outperforms DeepWide and YouTubeNet, the reason for which might be that DSSM models raw texts directly with word hashing.
				\item
					KPCNN performs best in all baselines.
					This is because KPCNN uses CNN to process input texts and can better extract the specific local patterns in sentences.
				\item
					Finally, compared with KPCNN, DKN can still have a $1.7\%$ AUC increase.
					We attribute the superiority of DKN to its two properties:
					1) DKN uses word-entity-aligned KCNN for sentence representation learning, which could better preserve the relatedness between words and entities;
					2) DKN uses an attention network to treat users' click history discriminatively, which better captures users' diverse reading interests.
			\end{itemize}
					
		Figure \ref{fig:result} presents the AUC score of DKN and baselines for additional ten test days.
		We can observe that the curve of DKN is consistently above baselines over ten days, which strongly proves the competitiveness of DKN.
		Moreover, the performance of DKN is also with low variance compared with baselines, which suggests that DKN is also robust and stable in practical application.
				
		\subsubsection{Comparison among DKN variants.}			
			Further, we compare among the variants of DKN with respect to the following four aspects to demonstrate the efficacy of the design of the DKN framework: the usage of knowledge, the choice of knowledge graph embedding method, the choice of transformation function, and the usage of an attention network.
			The results are shown in Table \ref{table:variants}, from which we can conclude that:
			\begin{itemize}
				\item
					The usage of entity embedding and contextual embedding can improve AUC by $1.3\%$ and $0.7\%$, respectively, and we can achieve even better performance by combining them together.
					This finding confirms the efficacy of using a knowledge graph in the DKN model.
				\item
					DKN+TransD outperforms other combinations.
					This is probably because, as presented in Section \ref{sec:kge}, TransD is the most complicated model among the four embedding methods, which is able to better capture non-linear relationships among the knowledge graph for news recommendation.
				\item
					DKN with mapping is better than DKN without mapping, and the non-linear function is superior to the linear one.
					The results prove that the transformation function can alleviate the heterogeneity between word and entity spaces by self learning, and the non-linear function can achieve better performance.
				\item
					The attention network brings a $1.7\%$ gain on F1 and $0.9\%$ gain on AUC for the DKN model.
					We will give a more intuitive demonstration on the attention network in the next subsection.
			\end{itemize}

	\subsection{Case Study}
	\label{sec:cs}        	
\begin{figure}[t]
			\centering
            \begin{subfigure}[b]{0.22\textwidth}
                \includegraphics[width=\textwidth]{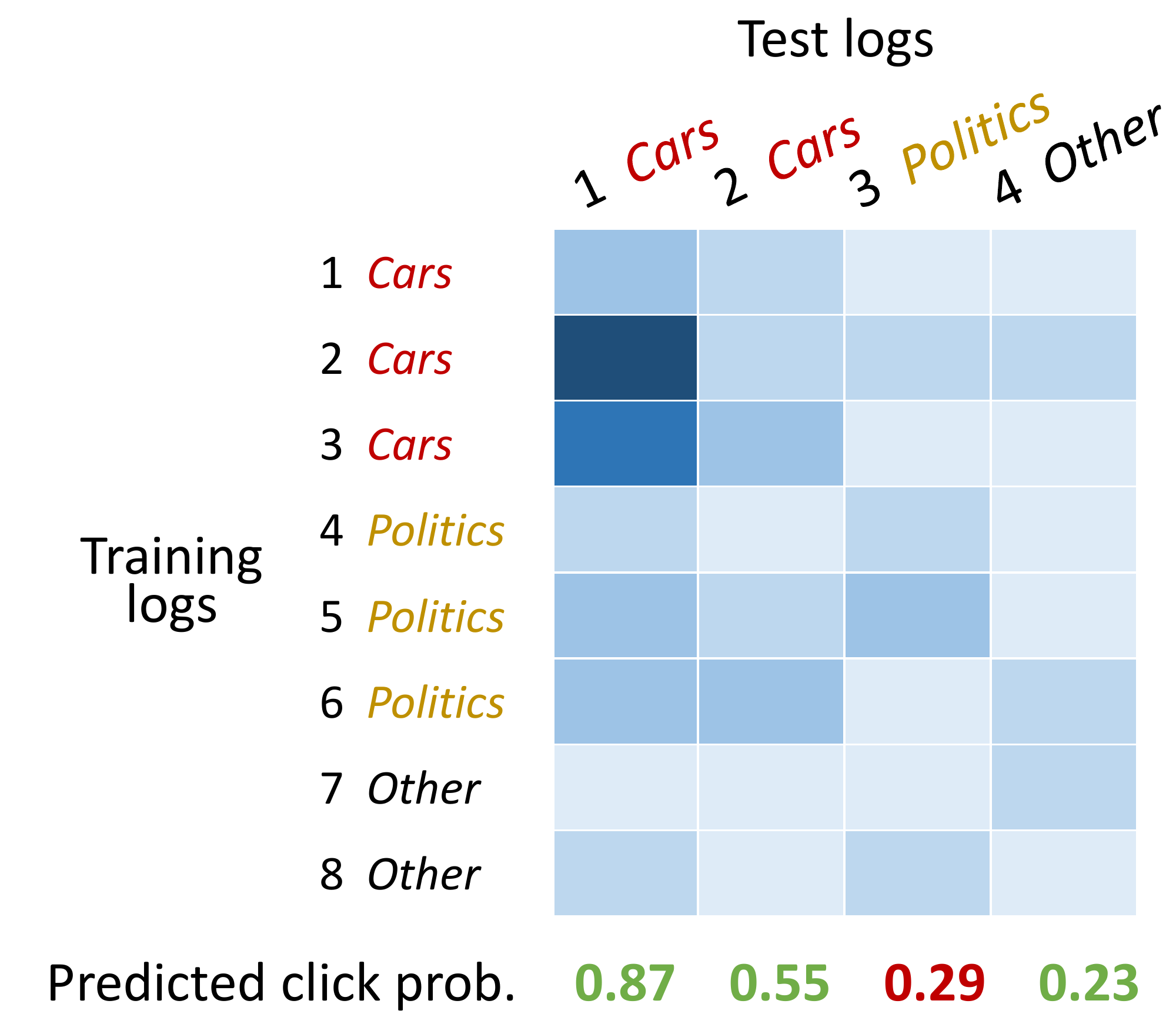}
                \caption{without knowledge graph}
                \label{fig:cs_a}
            \end{subfigure}
            \hfill
            \begin{subfigure}[b]{0.22\textwidth}
                \includegraphics[width=\textwidth]{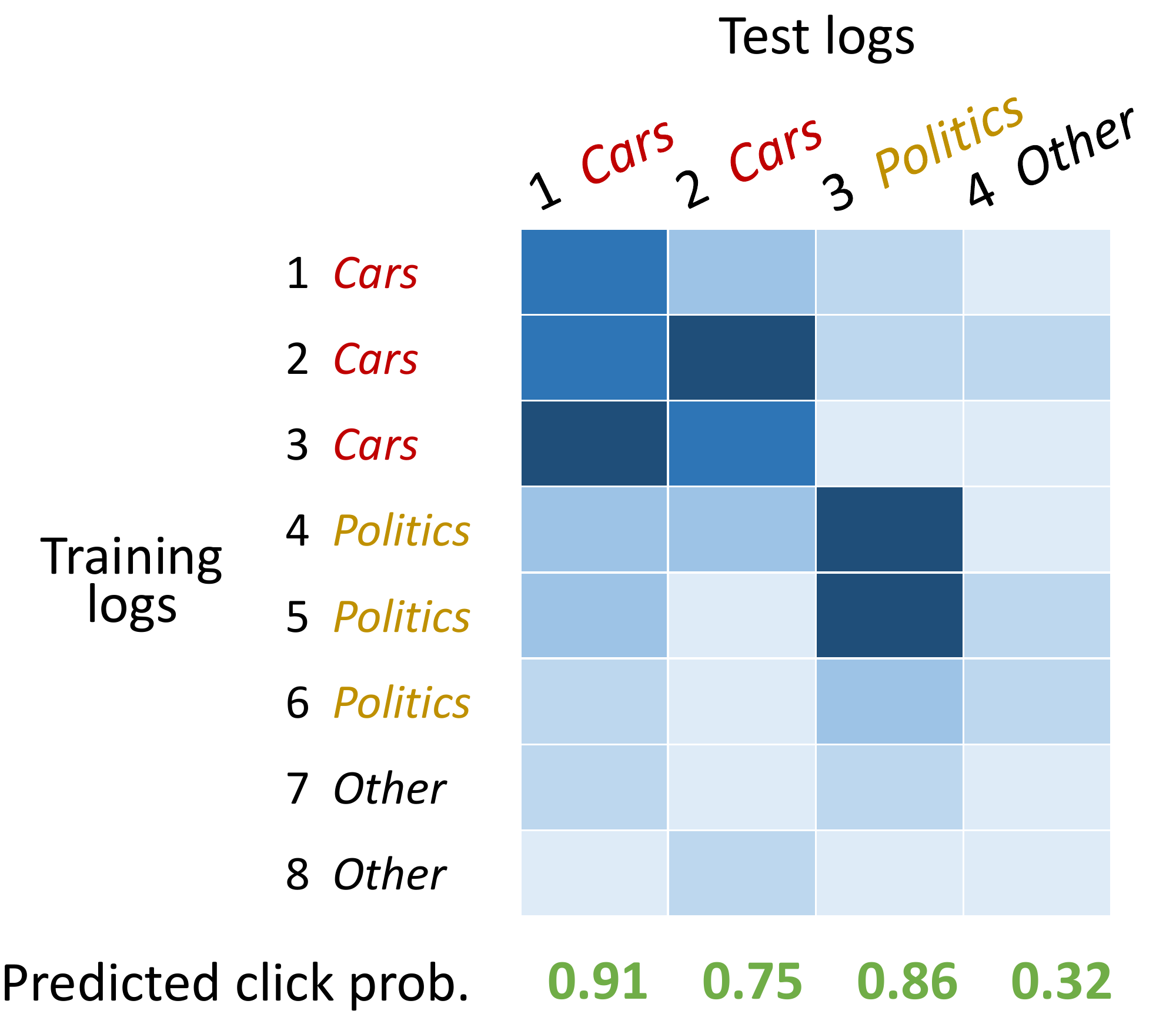}
                \caption{with knowledge graph}
                \label{fig:cs_b}
            \end{subfigure}
            \caption{Attention visualization for training logs and test logs for a randomly sampled user.}
            \label{fig:case_study}
            \vspace{-0.05in}
        \end{figure}

        	To intuitively demonstrate the efficacy of the usage of the knowledge graph as well as the the attention network, we randomly sample a user and extract all his logs from the training set and the test set (training logs with label 0 are omitted for simplicity).
        	As shown in Table \ref{table:case_study}, the clicked news clearly exhibits his points of interest: No. 1-3 concern cars and No. 4-6 concern politics (categories are not contained in the original dataset but manually tagged by us).
        	We use the whole training data to train DKN with full features and DKN without entity nor context embedding, then feed each possible pair of training logs and test logs of this user to the two trained models and obtain the output value of their attention networks.
        	The results are visualized in Figure \ref{fig:case_study}, in which the darker shade of blue indicates larger attention values.
        	From Figure \ref{fig:cs_a} we observe that, the first title in test logs gets high attention values with ``\textsf{Cars}'' in the training logs since they share the same word ``\textsf{Tesla}'', but the results for the second title are less satisfactory, since the second title shares no explicit word-similarity with any title in the training set, including No. 1-3.
        	The case is similar for the third title in test logs.
        	In contrast, in Figure \ref{fig:cs_b} we see that the attention network precisely captures the relatedness within the two categories ``\textsf{Cars}'' and ``\textsf{Politics}''.
        	This is because in the knowledge graph, ``\textsf{General Motors}'' and ``\textsf{Ford Inc.}'' share a large amount of context with ``\textsf{Tesla Inc.}'' and ``\textsf{Elon Musk}'', moreover, ``\textsf{Jeh Johnson}'' and ``\textsf{Russian}'' are also highly connected to ``\textsf{Donald Trump}''.
        	The difference in the response of the attention network also affects the final predicted results: DKN with knowledge graph (Figure \ref{fig:cs_b}) accurately predicts all the test logs, while DKN without knowledge graph (Figure \ref{fig:cs_a}) fails on the third one.

	\subsection{Parameter Sensitivity}
	\label{sec:ps}		
		DKN involves a number of hyper-parameters.
		In this subsection, we examine how different choices of hyper-parameters affect the performance of DKN.
		In the following experiments, expect for the parameter being tested, all other parameters are set as introduced in Section \ref{sec:es}.
		
		\subsubsection{Dimension of word embedding $d$ and dimension of entity embedding $k$.}
		We first investigate how the dimension of word embedding $d$ and dimension of entity embedding $k$ affect performance by testing all combinations of $d$ and $k$ in set $\{20, 50, 100, 200\}$.
		The results are shown in Figure \ref{fig:ps_a}, from which we can observe that, given dimension of entity embedding $k$, performance initially improves with the increase of dimension of word embedding $d$.
		This is because more bits in word embedding can encode more useful information of word semantics.
		However, the performance drops when $d$ further increases, as a too large $d$ (e.g., $d=200$) may introduce noises which mislead the subsequent prediction.
		The case is similar for $k$ when $d$ is given.
		
		\subsubsection{Window sizes of filters and the number of filters $m$.}
\begin{figure}[t]
			\centering
            \begin{subfigure}[b]{0.232\textwidth}
                \includegraphics[width=\textwidth]{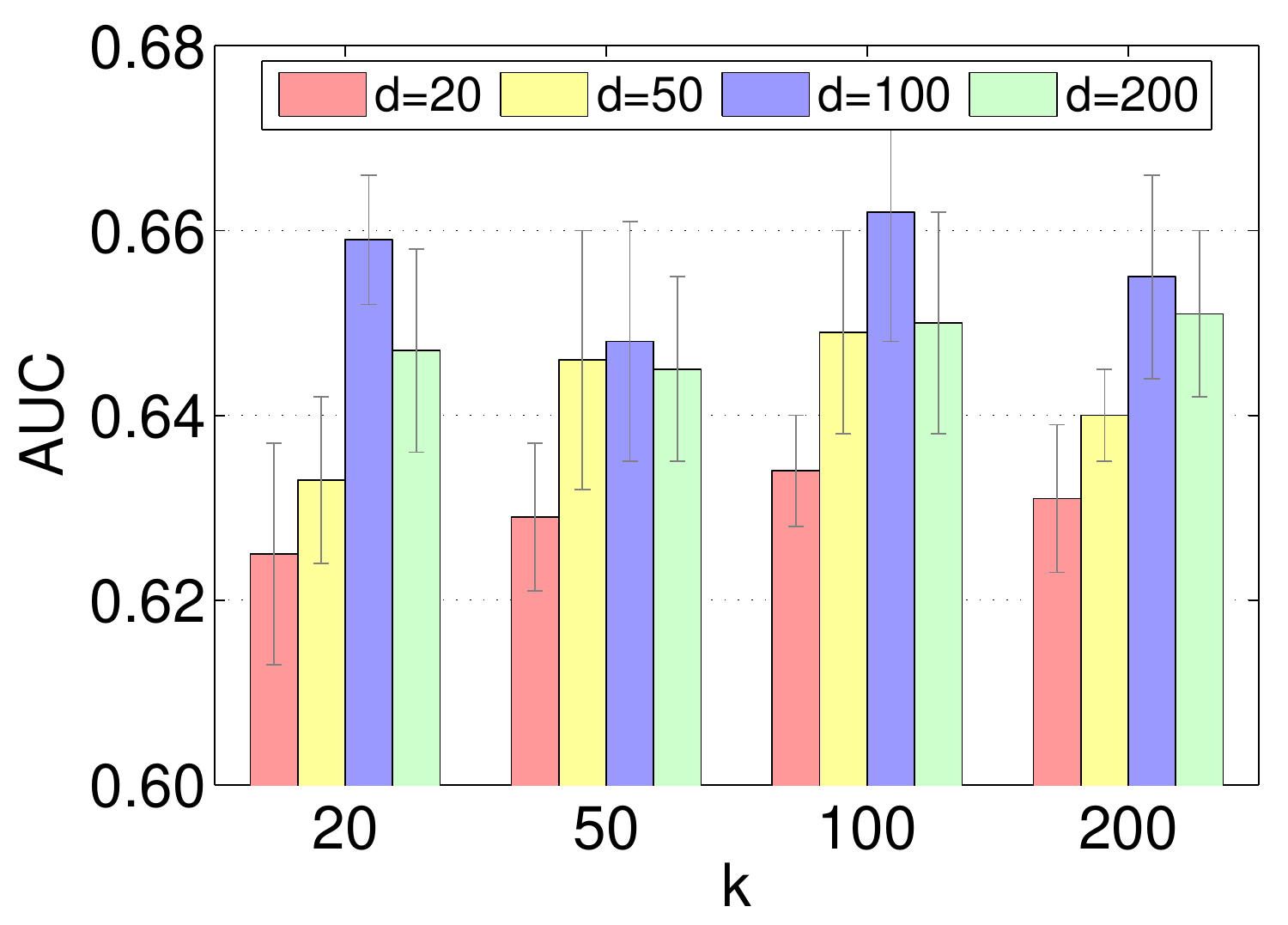}
                \caption{AUC score w.r.t dimension of entity embedding $k$ and dimension of word embedding $d$}
                \label{fig:ps_a}
            \end{subfigure}
            \hfill
            \begin{subfigure}[b]{0.232\textwidth}
                \includegraphics[width=\textwidth]{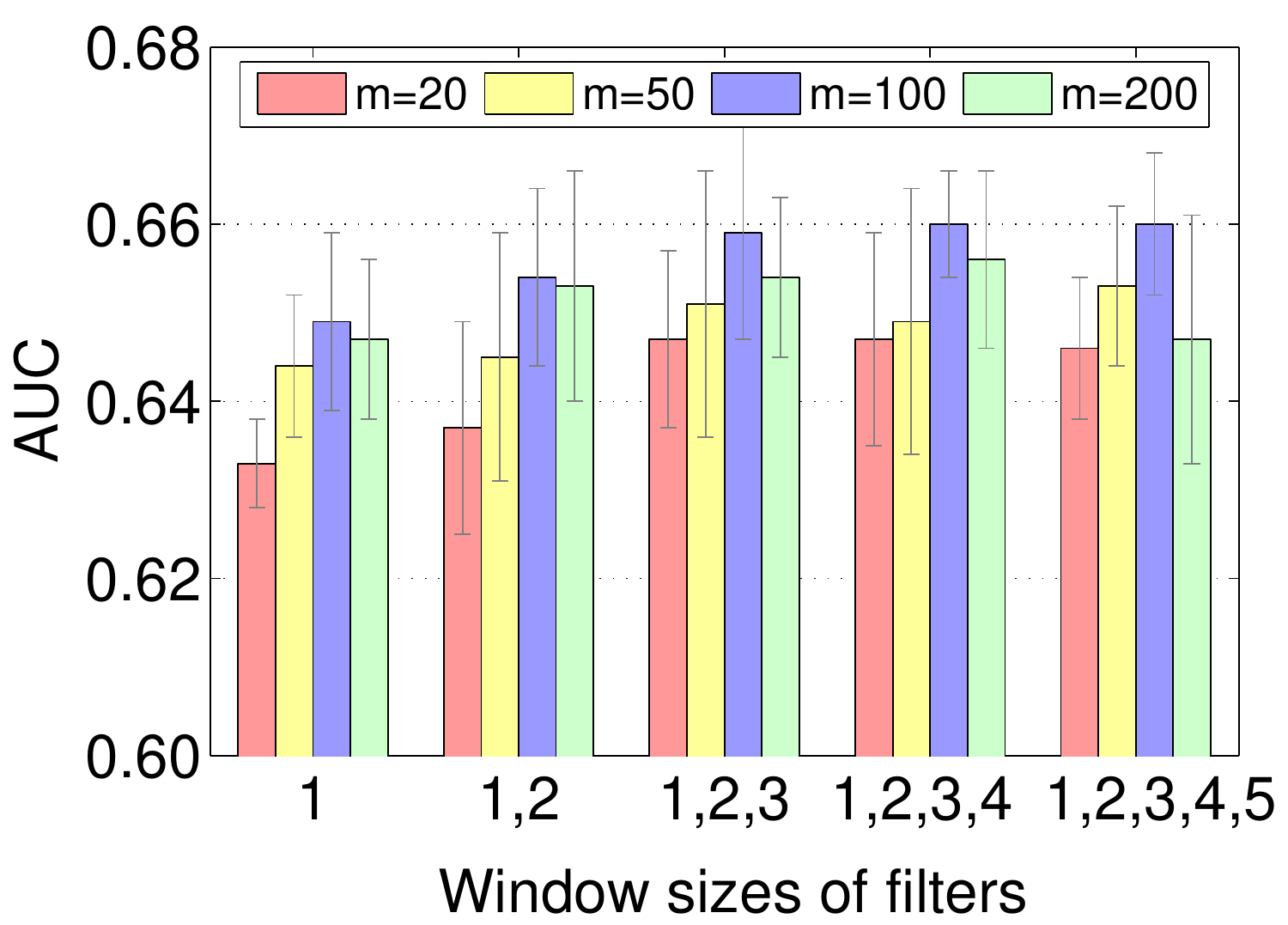}
                \caption{AUC score w.r.t window sizes of filters and the number of filters $m$}
                \label{fig:ps_b}
            \end{subfigure}
            \caption{Parameter sensitivity of DKN.}
            \label{fig:ps}
            \vspace{-0.05in}
        \end{figure}

		We further investigate the choice of windows sizes of filters and the number of filters for KCNN in the DKN model.
		As shown in Figure \ref{fig:ps_b}, given windows sizes, the AUC score generally increases as the number of filters $m$ gets larger, since more filters are able to capture more local patterns in input sentences and enhance model capability.
		However, the trend changes when $m$ is too large ($m = 200$) due to probable overfitting.
		Likewise, we can observe similar rules for window sizes given $m$: a small window size cannot capture long-distance patterns in sentences, while a too large window size may easily suffer from overfitting the noisy patterns.
		
\section{Related Work}
	\subsection{News Recommendation}
		News recommendation has previously been widely studied.
		Non-personalized news recommendation aims to model relatedness among news \cite{lv2011learning} or learn human editors' demonstration \cite{wang2017dynamic}.
		In personalized news recommendation, CF-based methods \cite{wang2011collaborative} often suffer from the cold-start problem since news items are substituted frequently.
		Therefore, a large amount of content-based or hybrid methods have been proposed \cite{kompan2010content, liu2010personalized, bansal2015content, phelan2009using, son2013location}.
		For example, \cite{phelan2009using} proposes a Bayesian method for predicting users' current news interests based on their click behavior, and \cite{son2013location} proposes an explicit localized sentiment analysis method for location-based news recommendation.
		Recently, researchers have also tried to combine other features into news recommendation, for example, contextual-bandit \cite{li2010contextual}, topic models \cite{luostarinen2013using}, and recurrent neural networks \cite{okura2017embedding}.
		The major difference between prior work and ours is that we use a knowledge graph to extract latent knowledge-level connections among news for better exploration in news recommendation.

	\subsection{Knowledge Graph}
		Knowledge graph representation aims to learn a low-dimensional vector for each entity and relation in the knowledge graph, while preserving the original graph structure.
		In addition to translation-based methods \cite{bordes2013translating, wang2014knowledge, lin2015learning, ji2015knowledge} used in DKN, researchers have also proposed many other models such as Structured Embedding \cite{bordes2011learning}, Latent Factor Model \cite{jenatton2012latent}, Neural Tensor Network \cite{socher2013reasoning} and GraphGAN \cite{wang2017graphgan}.
		Recently, the knowledge graph has also been used in many applications, such as movie recommendation\cite{zhang2016collaborative}, top-N recommendation \cite{palumbo2017entity2rec}, machine reading\cite{yang2017leveraging}, text classification\cite{wang2017combining} word embedding\cite{xu2014rc}, and question answering \cite{dong2015question}.
		To the best of our knowledge, this paper is the first work that proposes leveraging knowledge graph embedding in news recommendation.
		
	\subsection{Deep Recommender Systems}
		Recently, deep learning has been revolutionizing recommender systems and achieves better performance in many recommendation scenarios.
		Roughly speaking, deep recommender systems can be classified into two categories: using deep neural networks to process the raw features of users or items, or using deep neural networks to model the interaction among users and items.
		In addition to the aforementioned DSSM \cite{huang2013learning}, DeepWide \cite{cheng2016wide}, DeepFM \cite{guo2017deepfm}, YouTubeNet \cite{covington2016deep} and DMF \cite{xue2017deep}, other popular deep-learning-based recommender systems include Collaborative Deep Learning \cite{wang2015collaborative}, SHINE \cite{wang2017shine}, Multi-view Deep Learning \cite{elkahky2015multi}, and Neural Collaborative Filtering \cite{he2017neural}.
		The major difference between these methods and ours is that DKN specializes in news recommendation and could achieve better performance than other generic deep recommender systems.

\section{Conclusions}
	In this paper, we propose DKN, a deep knowledge-aware network that takes advantage of knowledge graph representation in news recommendation.
	DKN addresses three major challenges in news recommendation:
	1) Different from ID-based methods such as collaborative filtering, DKN is a content-based deep model for click-through rate prediction that are suitable for highly time-sensitive news.
	2) To make use of knowledge entities and common sense in news content, we design a KCNN module in DKN to jointly learn from semantic-level and knowledge-level representations of news.
	The multiple channels and alignment of words and entities enable KCNN to combine information from heterogeneous sources and maintain the correspondence of different embeddings for each word.
	3) To model the different impacts of a user's diverse historical interests on current candidate news, DKN uses an attention module to dynamically calculate a user's aggregated historical representation.
	We conduct extensive experiments on a dataset from Bing News.
	The results demonstrate the significant superiority of DKN compared with strong baselines, as well as the efficacy of the usage of knowledge entity embedding and the attention module.

\bibliographystyle{ACM-Reference-Format}
\bibliography{sigproc} 


\begin{thebibliography}{54}


\ifx \showCODEN    \undefined \def \showCODEN     #1{\unskip}     \fi
\ifx \showDOI      \undefined \def \showDOI       #1{#1}\fi
\ifx \showISBNx    \undefined \def \showISBNx     #1{\unskip}     \fi
\ifx \showISBNxiii \undefined \def \showISBNxiii  #1{\unskip}     \fi
\ifx \showISSN     \undefined \def \showISSN      #1{\unskip}     \fi
\ifx \showLCCN     \undefined \def \showLCCN      #1{\unskip}     \fi
\ifx \shownote     \undefined \def \shownote      #1{#1}          \fi
\ifx \showarticletitle \undefined \def \showarticletitle #1{#1}   \fi
\ifx \showURL      \undefined \def \showURL       {\relax}        \fi
\providecommand\bibfield[2]{#2}
\providecommand\bibinfo[2]{#2}
\providecommand\natexlab[1]{#1}
\providecommand\showeprint[2][]{arXiv:#2}

\bibitem[\protect\citeauthoryear{Agarwal and Chen}{Agarwal and Chen}{2009}]%
        {agarwal2009regression}
\bibfield{author}{\bibinfo{person}{Deepak Agarwal} {and}
  \bibinfo{person}{Bee-Chung Chen}.} \bibinfo{year}{2009}\natexlab{}.
\newblock \showarticletitle{Regression-based latent factor models}. In
  \bibinfo{booktitle}{{\em Proceedings of the 15th ACM SIGKDD international
  conference on Knowledge discovery and data mining}}. ACM,
  \bibinfo{pages}{19--28}.
\newblock


\bibitem[\protect\citeauthoryear{Bansal, Das, and Bhattacharyya}{Bansal
  et~al\mbox{.}}{2015}]%
        {bansal2015content}
\bibfield{author}{\bibinfo{person}{Trapit Bansal}, \bibinfo{person}{Mrinal
  Das}, {and} \bibinfo{person}{Chiranjib Bhattacharyya}.}
  \bibinfo{year}{2015}\natexlab{}.
\newblock \showarticletitle{Content driven user profiling for comment-worthy
  recommendations of news and blog articles}. In \bibinfo{booktitle}{{\em
  Proceedings of the 9th ACM Conference on Recommender Systems}}. ACM.
\newblock


\bibitem[\protect\citeauthoryear{Blei, Ng, and Jordan}{Blei
  et~al\mbox{.}}{2003}]%
        {blei2003latent}
\bibfield{author}{\bibinfo{person}{David~M Blei}, \bibinfo{person}{Andrew~Y
  Ng}, {and} \bibinfo{person}{Michael~I Jordan}.}
  \bibinfo{year}{2003}\natexlab{}.
\newblock \showarticletitle{Latent dirichlet allocation}.
\newblock \bibinfo{journal}{{\em Journal of machine Learning research\/}}
  \bibinfo{volume}{3}, \bibinfo{number}{Jan} (\bibinfo{year}{2003}),
  \bibinfo{pages}{993--1022}.
\newblock


\bibitem[\protect\citeauthoryear{Bordes, Usunier, Garcia-Duran, Weston, and
  Yakhnenko}{Bordes et~al\mbox{.}}{2013}]%
        {bordes2013translating}
\bibfield{author}{\bibinfo{person}{Antoine Bordes}, \bibinfo{person}{Nicolas
  Usunier}, \bibinfo{person}{Alberto Garcia-Duran}, \bibinfo{person}{Jason
  Weston}, {and} \bibinfo{person}{Oksana Yakhnenko}.}
  \bibinfo{year}{2013}\natexlab{}.
\newblock \showarticletitle{Translating embeddings for modeling
  multi-relational data}. In \bibinfo{booktitle}{{\em Advances in Neural
  Information Processing Systems}}. \bibinfo{pages}{2787--2795}.
\newblock


\bibitem[\protect\citeauthoryear{Bordes, Weston, Collobert, Bengio,
  et~al\mbox{.}}{Bordes et~al\mbox{.}}{2011}]%
        {bordes2011learning}
\bibfield{author}{\bibinfo{person}{Antoine Bordes}, \bibinfo{person}{Jason
  Weston}, \bibinfo{person}{Ronan Collobert}, \bibinfo{person}{Yoshua Bengio},
  {et~al\mbox{.}}} \bibinfo{year}{2011}\natexlab{}.
\newblock \showarticletitle{Learning Structured Embeddings of Knowledge
  Bases.}. In \bibinfo{booktitle}{{\em AAAI}}, Vol.~\bibinfo{volume}{6}.
  \bibinfo{pages}{6}.
\newblock


\bibitem[\protect\citeauthoryear{Cheng, Koc, Harmsen, Shaked, Chandra, Aradhye,
  Anderson, Corrado, Chai, Ispir, et~al\mbox{.}}{Cheng et~al\mbox{.}}{2016}]%
        {cheng2016wide}
\bibfield{author}{\bibinfo{person}{Heng-Tze Cheng}, \bibinfo{person}{Levent
  Koc}, \bibinfo{person}{Jeremiah Harmsen}, \bibinfo{person}{Tal Shaked},
  \bibinfo{person}{Tushar Chandra}, \bibinfo{person}{Hrishi Aradhye},
  \bibinfo{person}{Glen Anderson}, \bibinfo{person}{Greg Corrado},
  \bibinfo{person}{Wei Chai}, \bibinfo{person}{Mustafa Ispir}, {et~al\mbox{.}}}
  \bibinfo{year}{2016}\natexlab{}.
\newblock \showarticletitle{Wide \& deep learning for recommender systems}. In
  \bibinfo{booktitle}{{\em Proceedings of the 1st Workshop on Deep Learning for
  Recommender Systems}}. ACM, \bibinfo{pages}{7--10}.
\newblock


\bibitem[\protect\citeauthoryear{Conneau, Schwenk, Barrault, and Lecun}{Conneau
  et~al\mbox{.}}{2016}]%
        {conneau2016very}
\bibfield{author}{\bibinfo{person}{Alexis Conneau}, \bibinfo{person}{Holger
  Schwenk}, \bibinfo{person}{Lo{\"\i}c Barrault}, {and} \bibinfo{person}{Yann
  Lecun}.} \bibinfo{year}{2016}\natexlab{}.
\newblock \showarticletitle{Very deep convolutional networks for natural
  language processing}.
\newblock \bibinfo{journal}{{\em arXiv preprint arXiv:1606.01781\/}}
  (\bibinfo{year}{2016}).
\newblock


\bibitem[\protect\citeauthoryear{Covington, Adams, and Sargin}{Covington
  et~al\mbox{.}}{2016}]%
        {covington2016deep}
\bibfield{author}{\bibinfo{person}{Paul Covington}, \bibinfo{person}{Jay
  Adams}, {and} \bibinfo{person}{Emre Sargin}.}
  \bibinfo{year}{2016}\natexlab{}.
\newblock \showarticletitle{Deep neural networks for youtube recommendations}.
  In \bibinfo{booktitle}{{\em Proceedings of the 10th ACM Conference on
  Recommender Systems}}. ACM, \bibinfo{pages}{191--198}.
\newblock


\bibitem[\protect\citeauthoryear{Diao, Qiu, Wu, Smola, Jiang, and Wang}{Diao
  et~al\mbox{.}}{2014}]%
        {diao2014jointly}
\bibfield{author}{\bibinfo{person}{Qiming Diao}, \bibinfo{person}{Minghui Qiu},
  \bibinfo{person}{Chao-Yuan Wu}, \bibinfo{person}{Alexander~J Smola},
  \bibinfo{person}{Jing Jiang}, {and} \bibinfo{person}{Chong Wang}.}
  \bibinfo{year}{2014}\natexlab{}.
\newblock \showarticletitle{Jointly modeling aspects, ratings and sentiments
  for movie recommendation (jmars)}. In \bibinfo{booktitle}{{\em KDD}}. ACM,
  \bibinfo{pages}{193--202}.
\newblock


\bibitem[\protect\citeauthoryear{Dong, Wei, Zhou, and Xu}{Dong
  et~al\mbox{.}}{2015}]%
        {dong2015question}
\bibfield{author}{\bibinfo{person}{Li Dong}, \bibinfo{person}{Furu Wei},
  \bibinfo{person}{Ming Zhou}, {and} \bibinfo{person}{Ke Xu}.}
  \bibinfo{year}{2015}\natexlab{}.
\newblock \showarticletitle{Question Answering over Freebase with Multi-Column
  Convolutional Neural Networks.}. In \bibinfo{booktitle}{{\em ACL (1)}}.
\newblock


\bibitem[\protect\citeauthoryear{Elkahky, Song, and He}{Elkahky
  et~al\mbox{.}}{2015}]%
        {elkahky2015multi}
\bibfield{author}{\bibinfo{person}{Ali~Mamdouh Elkahky}, \bibinfo{person}{Yang
  Song}, {and} \bibinfo{person}{Xiaodong He}.} \bibinfo{year}{2015}\natexlab{}.
\newblock \showarticletitle{A multi-view deep learning approach for cross
  domain user modeling in recommendation systems}. In \bibinfo{booktitle}{{\em
  Proceedings of the 24th International Conference on World Wide Web}}.
  International World Wide Web Conferences Steering Committee,
  \bibinfo{pages}{278--288}.
\newblock


\bibitem[\protect\citeauthoryear{Fu, Liu, Ge, Yao, and Xiong}{Fu
  et~al\mbox{.}}{2014}]%
        {fu2014user}
\bibfield{author}{\bibinfo{person}{Yanjie Fu}, \bibinfo{person}{Bin Liu},
  \bibinfo{person}{Yong Ge}, \bibinfo{person}{Zijun Yao}, {and}
  \bibinfo{person}{Hui Xiong}.} \bibinfo{year}{2014}\natexlab{}.
\newblock \showarticletitle{User preference learning with multiple information
  fusion for restaurant recommendation}. In \bibinfo{booktitle}{{\em
  Proceedings of the 2014 SIAM International Conference on Data Mining}}. SIAM.
\newblock


\bibitem[\protect\citeauthoryear{Guo, Tang, Ye, Li, and He}{Guo
  et~al\mbox{.}}{2017}]%
        {guo2017deepfm}
\bibfield{author}{\bibinfo{person}{Huifeng Guo}, \bibinfo{person}{Ruiming
  Tang}, \bibinfo{person}{Yunming Ye}, \bibinfo{person}{Zhenguo Li}, {and}
  \bibinfo{person}{Xiuqiang He}.} \bibinfo{year}{2017}\natexlab{}.
\newblock \showarticletitle{DeepFM: A Factorization-Machine based Neural
  Network for CTR Prediction}. In \bibinfo{booktitle}{{\em Proceedings of the
  26th International Joint Conference on Artificial Intelligence}}.
\newblock


\bibitem[\protect\citeauthoryear{He, Liao, Zhang, Nie, Hu, and Chua}{He
  et~al\mbox{.}}{2017}]%
        {he2017neural}
\bibfield{author}{\bibinfo{person}{Xiangnan He}, \bibinfo{person}{Lizi Liao},
  \bibinfo{person}{Hanwang Zhang}, \bibinfo{person}{Liqiang Nie},
  \bibinfo{person}{Xia Hu}, {and} \bibinfo{person}{Tat-Seng Chua}.}
  \bibinfo{year}{2017}\natexlab{}.
\newblock \showarticletitle{Neural collaborative filtering}. In
  \bibinfo{booktitle}{{\em WWW}}. International World Wide Web Conferences
  Steering Committee, \bibinfo{pages}{173--182}.
\newblock


\bibitem[\protect\citeauthoryear{Hong and Fang}{Hong and Fang}{2015}]%
        {hong2015sentiment}
\bibfield{author}{\bibinfo{person}{James Hong} {and} \bibinfo{person}{Michael
  Fang}.} \bibinfo{year}{2015}\natexlab{}.
\newblock \bibinfo{booktitle}{{\em Sentiment analysis with deeply learned
  distributed representations of variable length texts}}.
\newblock \bibinfo{type}{{T}echnical {R}eport}. \bibinfo{institution}{Technical
  report, Stanford University}.
\newblock


\bibitem[\protect\citeauthoryear{Huang, He, Gao, Deng, Acero, and Heck}{Huang
  et~al\mbox{.}}{2013}]%
        {huang2013learning}
\bibfield{author}{\bibinfo{person}{Po-Sen Huang}, \bibinfo{person}{Xiaodong
  He}, \bibinfo{person}{Jianfeng Gao}, \bibinfo{person}{Li Deng},
  \bibinfo{person}{Alex Acero}, {and} \bibinfo{person}{Larry Heck}.}
  \bibinfo{year}{2013}\natexlab{}.
\newblock \showarticletitle{Learning deep structured semantic models for web
  search using clickthrough data}. In \bibinfo{booktitle}{{\em CIKM}}. ACM,
  \bibinfo{pages}{2333--2338}.
\newblock


\bibitem[\protect\citeauthoryear{Jenatton, Roux, Bordes, and
  Obozinski}{Jenatton et~al\mbox{.}}{2012}]%
        {jenatton2012latent}
\bibfield{author}{\bibinfo{person}{Rodolphe Jenatton},
  \bibinfo{person}{Nicolas~L Roux}, \bibinfo{person}{Antoine Bordes}, {and}
  \bibinfo{person}{Guillaume~R Obozinski}.} \bibinfo{year}{2012}\natexlab{}.
\newblock \showarticletitle{A latent factor model for highly multi-relational
  data}. In \bibinfo{booktitle}{{\em Advances in Neural Information Processing
  Systems}}. \bibinfo{pages}{3167--3175}.
\newblock


\bibitem[\protect\citeauthoryear{Ji, He, Xu, Liu, and Zhao}{Ji
  et~al\mbox{.}}{2015}]%
        {ji2015knowledge}
\bibfield{author}{\bibinfo{person}{Guoliang Ji}, \bibinfo{person}{Shizhu He},
  \bibinfo{person}{Liheng Xu}, \bibinfo{person}{Kang Liu}, {and}
  \bibinfo{person}{Jun Zhao}.} \bibinfo{year}{2015}\natexlab{}.
\newblock \showarticletitle{Knowledge Graph Embedding via Dynamic Mapping
  Matrix}. In \bibinfo{booktitle}{{\em ACL}}. \bibinfo{pages}{687--696}.
\newblock


\bibitem[\protect\citeauthoryear{Kalchbrenner, Grefenstette, and
  Blunsom}{Kalchbrenner et~al\mbox{.}}{2014}]%
        {kalchbrenner2014convolutional}
\bibfield{author}{\bibinfo{person}{Nal Kalchbrenner}, \bibinfo{person}{Edward
  Grefenstette}, {and} \bibinfo{person}{Phil Blunsom}.}
  \bibinfo{year}{2014}\natexlab{}.
\newblock \showarticletitle{A convolutional neural network for modelling
  sentences}.
\newblock \bibinfo{journal}{{\em arXiv preprint arXiv:1404.2188\/}}
  (\bibinfo{year}{2014}).
\newblock


\bibitem[\protect\citeauthoryear{Kim}{Kim}{2014}]%
        {kim2014convolutional}
\bibfield{author}{\bibinfo{person}{Yoon Kim}.} \bibinfo{year}{2014}\natexlab{}.
\newblock \showarticletitle{Convolutional neural networks for sentence
  classification}. In \bibinfo{booktitle}{{\em EMNLP}}.
\newblock


\bibitem[\protect\citeauthoryear{Kingma and Ba}{Kingma and Ba}{2014}]%
        {kingma2014adam}
\bibfield{author}{\bibinfo{person}{Diederik Kingma} {and}
  \bibinfo{person}{Jimmy Ba}.} \bibinfo{year}{2014}\natexlab{}.
\newblock \showarticletitle{Adam: A method for stochastic optimization}.
\newblock \bibinfo{journal}{{\em arXiv preprint arXiv:1412.6980\/}}
  (\bibinfo{year}{2014}).
\newblock


\bibitem[\protect\citeauthoryear{Kompan and Bielikov{\'a}}{Kompan and
  Bielikov{\'a}}{2010}]%
        {kompan2010content}
\bibfield{author}{\bibinfo{person}{Michal Kompan} {and}
  \bibinfo{person}{M{\'a}ria Bielikov{\'a}}.} \bibinfo{year}{2010}\natexlab{}.
\newblock \showarticletitle{Content-Based News Recommendation}. In
  \bibinfo{booktitle}{{\em EC-Web}}, Vol.~\bibinfo{volume}{61}. Springer,
  \bibinfo{pages}{61--72}.
\newblock


\bibitem[\protect\citeauthoryear{Krizhevsky, Sutskever, and Hinton}{Krizhevsky
  et~al\mbox{.}}{2012}]%
        {krizhevsky2012imagenet}
\bibfield{author}{\bibinfo{person}{Alex Krizhevsky}, \bibinfo{person}{Ilya
  Sutskever}, {and} \bibinfo{person}{Geoffrey~E Hinton}.}
  \bibinfo{year}{2012}\natexlab{}.
\newblock \showarticletitle{Imagenet classification with deep convolutional
  neural networks}. In \bibinfo{booktitle}{{\em Advances in neural information
  processing systems}}. \bibinfo{pages}{1097--1105}.
\newblock


\bibitem[\protect\citeauthoryear{Lai, Xu, Liu, and Zhao}{Lai
  et~al\mbox{.}}{2015}]%
        {lai2015recurrent}
\bibfield{author}{\bibinfo{person}{Siwei Lai}, \bibinfo{person}{Liheng Xu},
  \bibinfo{person}{Kang Liu}, {and} \bibinfo{person}{Jun Zhao}.}
  \bibinfo{year}{2015}\natexlab{}.
\newblock \showarticletitle{Recurrent Convolutional Neural Networks for Text
  Classification.}. In \bibinfo{booktitle}{{\em AAAI}},
  Vol.~\bibinfo{volume}{333}. \bibinfo{pages}{2267--2273}.
\newblock


\bibitem[\protect\citeauthoryear{Li, Chu, Langford, and Schapire}{Li
  et~al\mbox{.}}{2010}]%
        {li2010contextual}
\bibfield{author}{\bibinfo{person}{Lihong Li}, \bibinfo{person}{Wei Chu},
  \bibinfo{person}{John Langford}, {and} \bibinfo{person}{Robert~E Schapire}.}
  \bibinfo{year}{2010}\natexlab{}.
\newblock \showarticletitle{A contextual-bandit approach to personalized news
  article recommendation}. In \bibinfo{booktitle}{{\em Proceedings of the 19th
  international conference on World wide web}}. ACM, \bibinfo{pages}{661--670}.
\newblock


\bibitem[\protect\citeauthoryear{Lin, Liu, Sun, Liu, and Zhu}{Lin
  et~al\mbox{.}}{2015}]%
        {lin2015learning}
\bibfield{author}{\bibinfo{person}{Yankai Lin}, \bibinfo{person}{Zhiyuan Liu},
  \bibinfo{person}{Maosong Sun}, \bibinfo{person}{Yang Liu}, {and}
  \bibinfo{person}{Xuan Zhu}.} \bibinfo{year}{2015}\natexlab{}.
\newblock \showarticletitle{Learning Entity and Relation Embeddings for
  Knowledge Graph Completion}. In \bibinfo{booktitle}{{\em AAAI}}.
\newblock


\bibitem[\protect\citeauthoryear{Liu, Dolan, and Pedersen}{Liu
  et~al\mbox{.}}{2010}]%
        {liu2010personalized}
\bibfield{author}{\bibinfo{person}{Jiahui Liu}, \bibinfo{person}{Peter Dolan},
  {and} \bibinfo{person}{Elin~R{\o}nby Pedersen}.}
  \bibinfo{year}{2010}\natexlab{}.
\newblock \showarticletitle{Personalized news recommendation based on click
  behavior}. In \bibinfo{booktitle}{{\em Proceedings of the 15th international
  conference on Intelligent user interfaces}}. ACM, \bibinfo{pages}{31--40}.
\newblock


\bibitem[\protect\citeauthoryear{Luostarinen and Kohonen}{Luostarinen and
  Kohonen}{2013}]%
        {luostarinen2013using}
\bibfield{author}{\bibinfo{person}{Tapio Luostarinen} {and}
  \bibinfo{person}{Oskar Kohonen}.} \bibinfo{year}{2013}\natexlab{}.
\newblock \showarticletitle{Using topic models in content-based news
  recommender systems}. In \bibinfo{booktitle}{{\em Proceedings of the 19th
  Nordic Conference of Computational Linguistics}}. Link{\"o}ping University
  Electronic Press, \bibinfo{pages}{239--251}.
\newblock


\bibitem[\protect\citeauthoryear{Lv, Moon, Kolari, Zheng, Wang, and Chang}{Lv
  et~al\mbox{.}}{2011}]%
        {lv2011learning}
\bibfield{author}{\bibinfo{person}{Yuanhua Lv}, \bibinfo{person}{Taesup Moon},
  \bibinfo{person}{Pranam Kolari}, \bibinfo{person}{Zhaohui Zheng},
  \bibinfo{person}{Xuanhui Wang}, {and} \bibinfo{person}{Yi Chang}.}
  \bibinfo{year}{2011}\natexlab{}.
\newblock \showarticletitle{Learning to model relatedness for news
  recommendation}. In \bibinfo{booktitle}{{\em Proceedings of the 20th
  international conference on World wide web}}. ACM, \bibinfo{pages}{57--66}.
\newblock


\bibitem[\protect\citeauthoryear{Mikolov, Sutskever, Chen, Corrado, and
  Dean}{Mikolov et~al\mbox{.}}{2013}]%
        {mikolov2013distributed}
\bibfield{author}{\bibinfo{person}{Tomas Mikolov}, \bibinfo{person}{Ilya
  Sutskever}, \bibinfo{person}{Kai Chen}, \bibinfo{person}{Greg~S Corrado},
  {and} \bibinfo{person}{Jeff Dean}.} \bibinfo{year}{2013}\natexlab{}.
\newblock \showarticletitle{Distributed representations of words and phrases
  and their compositionality}. In \bibinfo{booktitle}{{\em Advances in neural
  information processing systems}}. \bibinfo{pages}{3111--3119}.
\newblock


\bibitem[\protect\citeauthoryear{Milne and Witten}{Milne and Witten}{2008}]%
        {milne2008learning}
\bibfield{author}{\bibinfo{person}{David Milne} {and} \bibinfo{person}{Ian~H
  Witten}.} \bibinfo{year}{2008}\natexlab{}.
\newblock \showarticletitle{Learning to link with wikipedia}. In
  \bibinfo{booktitle}{{\em CIKM}}. ACM, \bibinfo{pages}{509--518}.
\newblock


\bibitem[\protect\citeauthoryear{Okura, Tagami, Ono, and Tajima}{Okura
  et~al\mbox{.}}{2017}]%
        {okura2017embedding}
\bibfield{author}{\bibinfo{person}{Shumpei Okura}, \bibinfo{person}{Yukihiro
  Tagami}, \bibinfo{person}{Shingo Ono}, {and} \bibinfo{person}{Akira Tajima}.}
  \bibinfo{year}{2017}\natexlab{}.
\newblock \showarticletitle{Embedding-based News Recommendation for Millions of
  Users}. In \bibinfo{booktitle}{{\em KDD}}. ACM, \bibinfo{pages}{1933--1942}.
\newblock


\bibitem[\protect\citeauthoryear{Palumbo, Rizzo, and Troncy}{Palumbo
  et~al\mbox{.}}{2017}]%
        {palumbo2017entity2rec}
\bibfield{author}{\bibinfo{person}{Enrico Palumbo}, \bibinfo{person}{Giuseppe
  Rizzo}, {and} \bibinfo{person}{Rapha{\"e}l Troncy}.}
  \bibinfo{year}{2017}\natexlab{}.
\newblock \showarticletitle{entity2rec: Learning User-Item Relatedness from
  Knowledge Graphs for Top-N Item Recommendation}.
\newblock  (\bibinfo{year}{2017}).
\newblock


\bibitem[\protect\citeauthoryear{Phelan, McCarthy, and Smyth}{Phelan
  et~al\mbox{.}}{2009}]%
        {phelan2009using}
\bibfield{author}{\bibinfo{person}{Owen Phelan}, \bibinfo{person}{Kevin
  McCarthy}, {and} \bibinfo{person}{Barry Smyth}.}
  \bibinfo{year}{2009}\natexlab{}.
\newblock \showarticletitle{Using twitter to recommend real-time topical news}.
  In \bibinfo{booktitle}{{\em Proceedings of the third ACM conference on
  Recommender systems}}. ACM, \bibinfo{pages}{385--388}.
\newblock


\bibitem[\protect\citeauthoryear{Rendle}{Rendle}{2012}]%
        {rendle2012factorization}
\bibfield{author}{\bibinfo{person}{Steffen Rendle}.}
  \bibinfo{year}{2012}\natexlab{}.
\newblock \showarticletitle{Factorization machines with libfm}.
\newblock \bibinfo{journal}{{\em ACM Transactions on Intelligent Systems and
  Technology (TIST)\/}} \bibinfo{volume}{3}, \bibinfo{number}{3}
  (\bibinfo{year}{2012}), \bibinfo{pages}{57}.
\newblock


\bibitem[\protect\citeauthoryear{Sil and Yates}{Sil and Yates}{2013}]%
        {sil2013re}
\bibfield{author}{\bibinfo{person}{Avirup Sil} {and} \bibinfo{person}{Alexander
  Yates}.} \bibinfo{year}{2013}\natexlab{}.
\newblock \showarticletitle{Re-ranking for joint named-entity recognition and
  linking}. In \bibinfo{booktitle}{{\em Proceedings of the 22nd ACM
  international conference on Conference on information \& knowledge
  management}}. ACM, \bibinfo{pages}{2369--2374}.
\newblock


\bibitem[\protect\citeauthoryear{Socher, Chen, Manning, and Ng}{Socher
  et~al\mbox{.}}{2013a}]%
        {socher2013reasoning}
\bibfield{author}{\bibinfo{person}{Richard Socher}, \bibinfo{person}{Danqi
  Chen}, \bibinfo{person}{Christopher~D Manning}, {and} \bibinfo{person}{Andrew
  Ng}.} \bibinfo{year}{2013}\natexlab{a}.
\newblock \showarticletitle{Reasoning with neural tensor networks for knowledge
  base completion}. In \bibinfo{booktitle}{{\em Advances in neural information
  processing systems}}. \bibinfo{pages}{926--934}.
\newblock


\bibitem[\protect\citeauthoryear{Socher, Perelygin, Wu, Chuang, Manning, Ng,
  and Potts}{Socher et~al\mbox{.}}{2013b}]%
        {socher2013recursive}
\bibfield{author}{\bibinfo{person}{Richard Socher}, \bibinfo{person}{Alex
  Perelygin}, \bibinfo{person}{Jean Wu}, \bibinfo{person}{Jason Chuang},
  \bibinfo{person}{Christopher~D Manning}, \bibinfo{person}{Andrew Ng}, {and}
  \bibinfo{person}{Christopher Potts}.} \bibinfo{year}{2013}\natexlab{b}.
\newblock \showarticletitle{Recursive deep models for semantic compositionality
  over a sentiment treebank}. In \bibinfo{booktitle}{{\em Proceedings of the
  2013 conference on empirical methods in natural language processing}}.
  \bibinfo{pages}{1631--1642}.
\newblock


\bibitem[\protect\citeauthoryear{Son, Kim, Park, et~al\mbox{.}}{Son
  et~al\mbox{.}}{2013}]%
        {son2013location}
\bibfield{author}{\bibinfo{person}{Jeong-Woo Son}, \bibinfo{person}{A Kim},
  \bibinfo{person}{Seong-Bae Park}, {et~al\mbox{.}}}
  \bibinfo{year}{2013}\natexlab{}.
\newblock \showarticletitle{A location-based news article recommendation with
  explicit localized semantic analysis}. In \bibinfo{booktitle}{{\em
  Proceedings of the 36th international ACM SIGIR conference on Research and
  development in information retrieval}}. ACM, \bibinfo{pages}{293--302}.
\newblock


\bibitem[\protect\citeauthoryear{Tai, Socher, and Manning}{Tai
  et~al\mbox{.}}{2015}]%
        {tai2015improved}
\bibfield{author}{\bibinfo{person}{Kai~Sheng Tai}, \bibinfo{person}{Richard
  Socher}, {and} \bibinfo{person}{Christopher~D Manning}.}
  \bibinfo{year}{2015}\natexlab{}.
\newblock \showarticletitle{Improved semantic representations from
  tree-structured long short-term memory networks}.
\newblock \bibinfo{journal}{{\em arXiv preprint arXiv:1503.00075\/}}
  (\bibinfo{year}{2015}).
\newblock


\bibitem[\protect\citeauthoryear{Wang and Blei}{Wang and Blei}{2011}]%
        {wang2011collaborative}
\bibfield{author}{\bibinfo{person}{Chong Wang} {and} \bibinfo{person}{David~M
  Blei}.} \bibinfo{year}{2011}\natexlab{}.
\newblock \showarticletitle{Collaborative topic modeling for recommending
  scientific articles}. In \bibinfo{booktitle}{{\em Proceedings of the 17th ACM
  SIGKDD international conference on Knowledge discovery and data mining}}.
  ACM, \bibinfo{pages}{448--456}.
\newblock


\bibitem[\protect\citeauthoryear{Wang, Wang, Wang, Zhao, Zhang, Zhang, Xie, and
  Guo}{Wang et~al\mbox{.}}{2018a}]%
        {wang2017graphgan}
\bibfield{author}{\bibinfo{person}{Hongwei Wang}, \bibinfo{person}{Jia Wang},
  \bibinfo{person}{Jialin Wang}, \bibinfo{person}{Miao Zhao},
  \bibinfo{person}{Weinan Zhang}, \bibinfo{person}{Fuzheng Zhang},
  \bibinfo{person}{Xing Xie}, {and} \bibinfo{person}{Minyi Guo}.}
  \bibinfo{year}{2018}\natexlab{a}.
\newblock \showarticletitle{GraphGAN: Graph Representation Learning with
  Generative Adversarial Nets}. In \bibinfo{booktitle}{{\em AAAI}}.
\newblock


\bibitem[\protect\citeauthoryear{Wang, Wang, Zhao, Cao, and Guo}{Wang
  et~al\mbox{.}}{2017b}]%
        {wang2017joint}
\bibfield{author}{\bibinfo{person}{Hongwei Wang}, \bibinfo{person}{Jia Wang},
  \bibinfo{person}{Miao Zhao}, \bibinfo{person}{Jiannong Cao}, {and}
  \bibinfo{person}{Minyi Guo}.} \bibinfo{year}{2017}\natexlab{b}.
\newblock \showarticletitle{Joint-Topic-Semantic-aware Social Recommendation
  for Online Voting}. In \bibinfo{booktitle}{{\em Proceedings of the 26th ACM
  International Conference on Conference on Information and Knowledge
  Management}}. ACM, \bibinfo{pages}{347--356}.
\newblock


\bibitem[\protect\citeauthoryear{Wang, Wang, and Yeung}{Wang
  et~al\mbox{.}}{2015}]%
        {wang2015collaborative}
\bibfield{author}{\bibinfo{person}{Hao Wang}, \bibinfo{person}{Naiyan Wang},
  {and} \bibinfo{person}{Dit-Yan Yeung}.} \bibinfo{year}{2015}\natexlab{}.
\newblock \showarticletitle{Collaborative deep learning for recommender
  systems}. In \bibinfo{booktitle}{{\em Proceedings of the 21th ACM SIGKDD
  International Conference on Knowledge Discovery and Data Mining}}. ACM,
  \bibinfo{pages}{1235--1244}.
\newblock


\bibitem[\protect\citeauthoryear{Wang, Zhang, Hou, Xie, Guo, and Liu}{Wang
  et~al\mbox{.}}{2018b}]%
        {wang2017shine}
\bibfield{author}{\bibinfo{person}{Hongwei Wang}, \bibinfo{person}{Fuzheng
  Zhang}, \bibinfo{person}{Min Hou}, \bibinfo{person}{Xing Xie},
  \bibinfo{person}{Minyi Guo}, {and} \bibinfo{person}{Qi Liu}.}
  \bibinfo{year}{2018}\natexlab{b}.
\newblock \showarticletitle{Shine: Signed heterogeneous information network
  embedding for sentiment link prediction}. In \bibinfo{booktitle}{{\em WSDM}}.
\newblock


\bibitem[\protect\citeauthoryear{Wang, Wang, Zhang, and Yan}{Wang
  et~al\mbox{.}}{2017a}]%
        {wang2017combining}
\bibfield{author}{\bibinfo{person}{Jin Wang}, \bibinfo{person}{Zhongyuan Wang},
  \bibinfo{person}{Dawei Zhang}, {and} \bibinfo{person}{Jun Yan}.}
  \bibinfo{year}{2017}\natexlab{a}.
\newblock \showarticletitle{Combining Knowledge with Deep Convolutional Neural
  Networks for Short Text Classification}. In \bibinfo{booktitle}{{\em
  Proceedings of the International Joint Conference on Artificial
  Intelligence}}.
\newblock


\bibitem[\protect\citeauthoryear{Wang, Yu, Ren, Tao, Zhang, Yu, and Wang}{Wang
  et~al\mbox{.}}{2017c}]%
        {wang2017dynamic}
\bibfield{author}{\bibinfo{person}{Xuejian Wang}, \bibinfo{person}{Lantao Yu},
  \bibinfo{person}{Kan Ren}, \bibinfo{person}{Guanyu Tao},
  \bibinfo{person}{Weinan Zhang}, \bibinfo{person}{Yong Yu}, {and}
  \bibinfo{person}{Jun Wang}.} \bibinfo{year}{2017}\natexlab{c}.
\newblock \showarticletitle{Dynamic Attention Deep Model for Article
  Recommendation by Learning Human Editors' Demonstration}. In
  \bibinfo{booktitle}{{\em KDD}}. ACM.
\newblock


\bibitem[\protect\citeauthoryear{Wang, Zhang, Feng, and Chen}{Wang
  et~al\mbox{.}}{2014}]%
        {wang2014knowledge}
\bibfield{author}{\bibinfo{person}{Zhen Wang}, \bibinfo{person}{Jianwen Zhang},
  \bibinfo{person}{Jianlin Feng}, {and} \bibinfo{person}{Zheng Chen}.}
  \bibinfo{year}{2014}\natexlab{}.
\newblock \showarticletitle{Knowledge Graph Embedding by Translating on
  Hyperplanes}. In \bibinfo{booktitle}{{\em AAAI}}.
  \bibinfo{pages}{1112--1119}.
\newblock


\bibitem[\protect\citeauthoryear{Xu, Bai, Bian, Gao, Wang, Liu, and Liu}{Xu
  et~al\mbox{.}}{2014}]%
        {xu2014rc}
\bibfield{author}{\bibinfo{person}{Chang Xu}, \bibinfo{person}{Yalong Bai},
  \bibinfo{person}{Jiang Bian}, \bibinfo{person}{Bin Gao},
  \bibinfo{person}{Gang Wang}, \bibinfo{person}{Xiaoguang Liu}, {and}
  \bibinfo{person}{Tie-Yan Liu}.} \bibinfo{year}{2014}\natexlab{}.
\newblock \showarticletitle{Rc-net: A general framework for incorporating
  knowledge into word representations}. In \bibinfo{booktitle}{{\em Proceedings
  of the 23rd ACM International Conference on Conference on Information and
  Knowledge Management}}. ACM, \bibinfo{pages}{1219--1228}.
\newblock


\bibitem[\protect\citeauthoryear{Xue, Dai, Zhang, Huang, and Chen}{Xue
  et~al\mbox{.}}{2017}]%
        {xue2017deep}
\bibfield{author}{\bibinfo{person}{Hong-Jian Xue}, \bibinfo{person}{Xin-Yu
  Dai}, \bibinfo{person}{Jianbing Zhang}, \bibinfo{person}{Shujian Huang},
  {and} \bibinfo{person}{Jiajun Chen}.} \bibinfo{year}{2017}\natexlab{}.
\newblock \showarticletitle{Deep Matrix Factorization Models for Recommender
  Systems}. In \bibinfo{booktitle}{{\em Proceedings of the 26th International
  Joint Conference on Artificial Intelligence}}.
\newblock


\bibitem[\protect\citeauthoryear{Yang and Mitchell}{Yang and Mitchell}{2017}]%
        {yang2017leveraging}
\bibfield{author}{\bibinfo{person}{Bishan Yang} {and} \bibinfo{person}{Tom
  Mitchell}.} \bibinfo{year}{2017}\natexlab{}.
\newblock \showarticletitle{Leveraging knowledge bases in lstms for improving
  machine reading}. In \bibinfo{booktitle}{{\em Proceedings of the 55th Annual
  Meeting of the Association for Computational Linguistics (Volume 1: Long
  Papers)}}, Vol.~\bibinfo{volume}{1}.
\newblock


\bibitem[\protect\citeauthoryear{Zhang, Yuan, Lian, Xie, and Ma}{Zhang
  et~al\mbox{.}}{2016}]%
        {zhang2016collaborative}
\bibfield{author}{\bibinfo{person}{Fuzheng Zhang},
  \bibinfo{person}{Nicholas~Jing Yuan}, \bibinfo{person}{Defu Lian},
  \bibinfo{person}{Xing Xie}, {and} \bibinfo{person}{Wei-Ying Ma}.}
  \bibinfo{year}{2016}\natexlab{}.
\newblock \showarticletitle{Collaborative knowledge base embedding for
  recommender systems}. In \bibinfo{booktitle}{{\em KDD}}. ACM,
  \bibinfo{pages}{353--362}.
\newblock


\bibitem[\protect\citeauthoryear{Zhang, Zhao, and LeCun}{Zhang
  et~al\mbox{.}}{2015}]%
        {zhang2015character}
\bibfield{author}{\bibinfo{person}{Xiang Zhang}, \bibinfo{person}{Junbo Zhao},
  {and} \bibinfo{person}{Yann LeCun}.} \bibinfo{year}{2015}\natexlab{}.
\newblock \showarticletitle{Character-level convolutional networks for text
  classification}. In \bibinfo{booktitle}{{\em NIPS}}.
  \bibinfo{pages}{649--657}.
\newblock


\bibitem[\protect\citeauthoryear{Zhou, Song, Zhu, Ma, Yan, Dai, Zhu, Jin, Li,
  and Gai}{Zhou et~al\mbox{.}}{2017}]%
        {zhou2017deep}
\bibfield{author}{\bibinfo{person}{Guorui Zhou}, \bibinfo{person}{Chengru
  Song}, \bibinfo{person}{Xiaoqiang Zhu}, \bibinfo{person}{Xiao Ma},
  \bibinfo{person}{Yanghui Yan}, \bibinfo{person}{Xingya Dai},
  \bibinfo{person}{Han Zhu}, \bibinfo{person}{Junqi Jin}, \bibinfo{person}{Han
  Li}, {and} \bibinfo{person}{Kun Gai}.} \bibinfo{year}{2017}\natexlab{}.
\newblock \showarticletitle{Deep Interest Network for Click-Through Rate
  Prediction}.
\newblock \bibinfo{journal}{{\em arXiv preprint arXiv:1706.06978\/}}
  (\bibinfo{year}{2017}).
\newblock


\end{thebibliography}

\end{document}